\pdfoutput=1

\documentclass[11pt]{article}
\usepackage{algorithm}
\usepackage{algpseudocode}
\usepackage{svg}
\usepackage[]{ACL2023}
\usepackage{graphicx}
\usepackage{times}
\usepackage{latexsym}
\usepackage{multirow}
\usepackage[T1]{fontenc}

\usepackage[utf8]{inputenc}
\usepackage{wrapfig}
\usepackage{microtype}                         
\usepackage{booktabs}
\usepackage{inconsolata}
\usepackage{amsmath}
\usepackage{amssymb}

\usepackage{rotating}

\title{VillagerAgent: A Graph-Based Multi-Agent Framework for Coordinating Complex Task Dependencies in Minecraft}

\author{Yubo Dong, Xukun Zhu, Zhengzhe Pan, Linchao Zhu$^\dag$, Yi Yang\\ 
        ReLER, CCAI, Zhejiang University\\ 
        $\dag$ Corresponding author} 

\begin{document}
\maketitle

\begin{abstract}

In this paper, we aim to evaluate multi-agent systems against complex dependencies, including spatial, causal, and temporal constraints. 
First, we construct a new benchmark, named \textbf{VillagerBench}, within the Minecraft environment.
VillagerBench comprises diverse tasks 
crafted to test various aspects of multi-agent collaboration, from workload distribution to dynamic adaptation and synchronized task execution.
Second, we introduce a Directed Acyclic Graph Multi-Agent Framework (\textbf{VillagerAgent}) to resolve complex inter-agent dependencies and enhance collaborative efficiency. This solution incorporates a task decomposer that creates a directed acyclic graph (DAG) for structured task management, an agent controller for task distribution, and a state manager for tracking environmental and agent data.
Our empirical evaluation on VillagerBench demonstrates that VillagerAgent outperforms the existing AgentVerse model, reducing hallucinations and improving task decomposition efficacy. The results underscore VillagerAgent's potential in advancing multi-agent collaboration, offering a scalable and generalizable solution in dynamic environments.
Source code is open-source on GitHub.
\footnote{\url{https://github.com/cnsdqd-dyb/VillagerAgent}}
\end{abstract}



\section{Introduction}


\noindent Multi-agent collaboration using LLM is a challenging research topic that aims to enable multiple autonomous agents to coordinate their actions and achieve a common goal \cite{wang2023survey,xi2023rise,qian2023experiential,qian2023communicative,xie2023openagents,wu2023autogen}. The collaboration process requires communication, planning, and reasoning among multiple intelligent agents. It has many applications in domains such as robotics, gaming \cite{wang2023voyager}, and social simulation \cite{li2023camel}. 

\begin{figure}[tp] 
  \centering 
  \includegraphics[width=\linewidth]{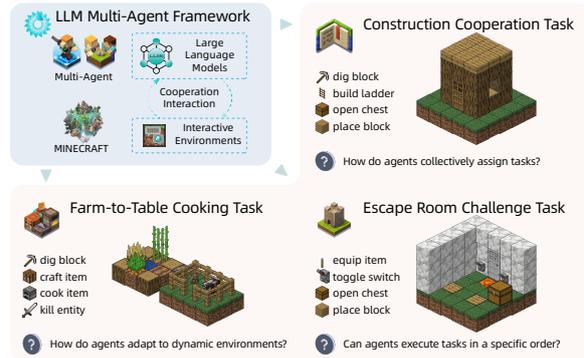} 
  \caption{Minecraft Multi-Agent Benchmark (VillagerBench) is the first multi-scenario benchmark designed to evaluate the cooperative capabilities of multi-agent systems within the real-world context of Minecraft.} 
  \label{fig:m2ab} 
\end{figure}

\noindent There are increasing interests in developing multi-agent systems using LLMs.
MindAgent introduces the CuisineWorld gaming scenario as a benchmark, utilizing the Collaboration Score (CoS) to measure the efficiency of collaboration~\cite{gong2023mindagent}. AgentVerse organizes its framework into four essential stages: Expert Recruitment, Collaborative Decision-Making, Action Execution, and Evaluation, thereby effectively deploying multi-agent groups that outperform a single agent~\cite{chen2023agentverse}.
MetaGPT, on the other hand, employs an assembly line approach, designating specific roles to agents and efficiently breaking down complex tasks into subtasks involving many agents working together~\cite{hong2023metagpt}.
However, these multi-agent collaboration models either tend to restrict agents to parallel-executable subtasks each round, even when unnecessary or bind them to a fixed pipeline and task stage, overlooking complex task dependencies. This may cause issues for tasks that need both sequential and parallel execution, thus limiting model generality and scalability~\cite{gong2023mindagent,chen2023agentverse,hong2023metagpt}.


\noindent In this paper, we focus on multi-agent collaboration for problem-solving with complex dependencies. These dependencies can be of different types, such as spatial dependencies that constrain the locations of the sub-tasks, causal dependencies that affect the availability and effects of the sub-tasks, and temporal dependencies that impose constraints on the timing of the sub-tasks. It is crucial to understand and manage these dependencies for effective multi-agent collaboration, enabling the agents to reason about the long-term consequences of their actions and avoid potential conflicts.

\noindent First, we introduce VillagerBench, a new multi-agent benchmark in the Minecraft environment designed for the evaluation of complex dependencies (Figure \ref{fig:display}).
Some of the multi-agent research is being tested within the Overcooked-AI \cite{carroll2020utility}. Nevertheless, due to limitations in the number of agents, scenario flexibility, and task diversity, there is a desire for more comprehensive frameworks to test multi-agent cooperation. 
Inspired by Voyager~\cite{wang2023voyager}, GITM~\cite{zhu2023ghost}, and MindAgent~\cite{gong2023mindagent}, we construct a multi-agent and multi-task evaluation framework with greater degrees of freedom using Minecraft.
Minecraft offers a rich and diverse set of tasks that can be used to benchmark and evaluate multi-agent systems, such as building and farming. It allows players to explore dynamic environments that pose various challenges for multi-agent collaboration, such as resource allocation, task decomposition, and coordination. 
Specifically, we introduce three tasks, i.e., Construction Cooperation, Farm-to-Table Cooking and Escape Room Challenge. The Construction Cooperation task tests agents' aptitude for understanding task requirements and orchestrating team workload, focusing on the evaluation of spatial dependencies in multi-agent collaboration. The Farm-to-Table Cooking task assesses their agility in adapting to fluctuating environmental conditions, aiming to solve complex causal dependencies. 
The Escape Room Challenge task tests agents on their ability to execute tasks both sequentially and in parallel, requiring the reasoning of temporal dependencies and the ability to synchronize actions.

\noindent Second, we introduce a Directed Acyclic Graph Multi-Agent framework (VillagerAgent) to tackle complex dependencies in multi-agent collaborations.
Each subtask is represented as a graph node in the DAG. We dynamically adjust the graph structure and the agent roles according to the environment and the agent states. 
VillagerAgent consists of task decomposer, agent controller, state manager and base agents. The Task Decomposer generate a Directed Acyclic Graph (DAG) of subtask nodes each round, while the Agent Controller oversees the assignment of these subtasks to the Base Agents for execution and self-reflection. Meanwhile, the State Manager is responsible for maintaining the status information of both the environment and the agents. 

\noindent We quantitatively evaluate our method on VillagerBench.
We demonstrate the superior performance of VillagerAgent over AgentVerse~\cite{chen2023agentverse} by fewer hallucinations and enhancing the effectiveness of task decomposition.

\label{VillagerAgent}
\begin{figure*}[htbp] 
  \centering 
  \includegraphics[width=\textwidth]{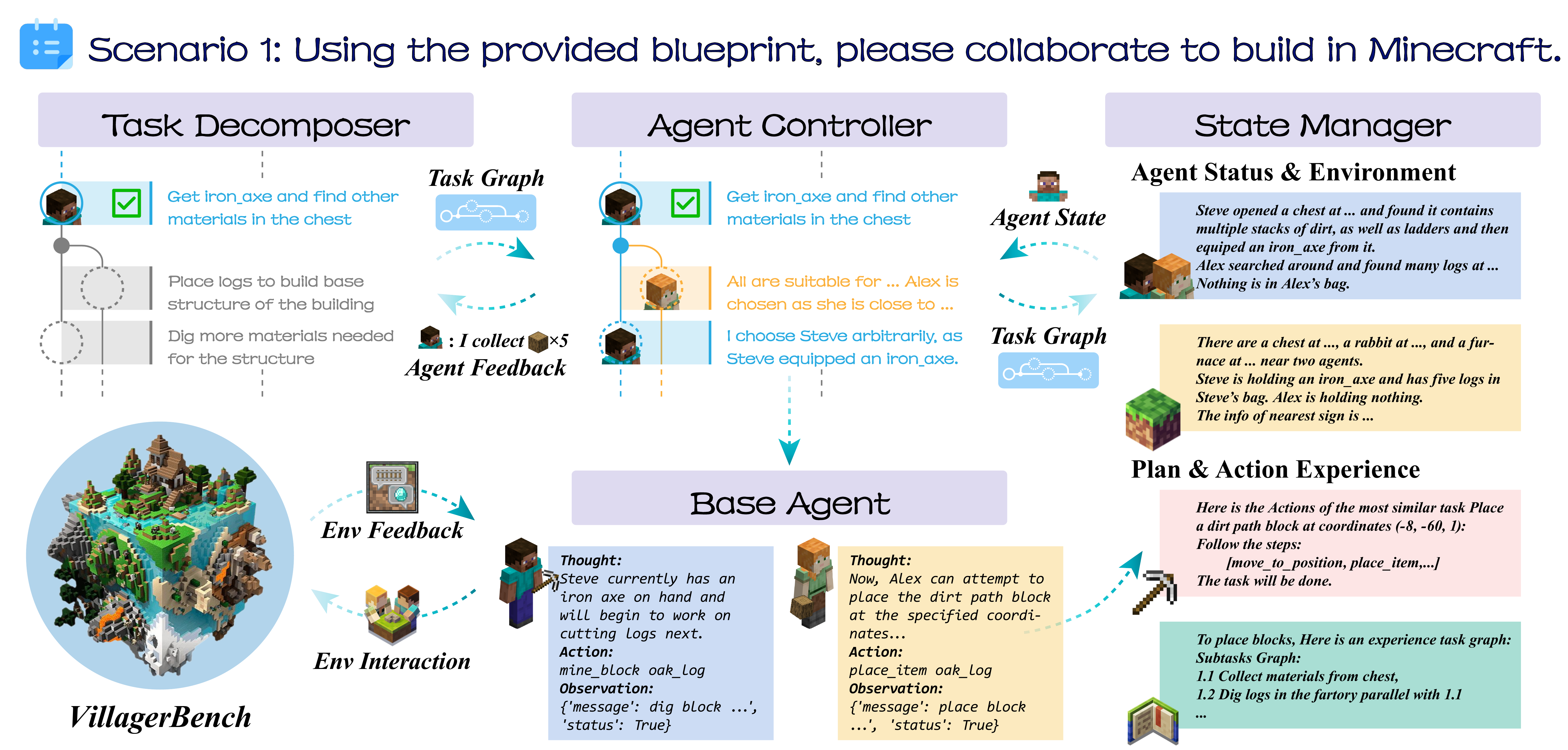} 
  \caption{Overview of the VillagerAgent framework. Our framework acts as the central architecture for individual agents, enhancing their collaborative capabilities. Featuring a Task Decomposer that generates subtask DAGs, an Agent Controller for task assignment, a State Manager for status updating, and Base Agents for task execution and self-assessment.} 
  \label{fig:flowchart} 
\end{figure*}

\section{VillagerBench Design}

\noindent Our VillagerBench uses Mineflayer~\cite{mineflayer} to establish Agent APIs, offering a platform to examine cooperative behaviors in multi-agent systems via tasks such as construction, cooking, and escape room challenges (Figure~\ref{fig:m2ab}).

\noindent We evaluate multi-agent systems powered by LLMs using three key metrics: \textbf{Completion (C)} that measures the average task completion rate; \textbf{Efficiency (E)} that assesses the speed of task execution and the utilization of resources; and \textbf{Balance (B)} that examines the distribution of workload among agents, with higher values indicating a more equitable assignment of tasks. Further details can be found in Appendix~\ref{sec:Metrics}.

\paragraph{Construction Cooperation Task: Interpretation and Allocation.}
\noindent Construction Cooperation task is centered around the agents' proficiency in interpreting detailed task documents and efficiently allocating the workload among team members. This task necessitates a high level of comprehension and coordination, as agents must parse the project specifications and judiciously assign sub-tasks to optimize collective performance.

\noindent Agents are provided with textual architectural blueprints that specify the positions and orientations of blocks required for construction tasks. Building materials are supplied in chests or at a material factory, where agents must mine and transport them to the building site. Further details can be found in Appendix~\ref{construction illustrate}.

\paragraph{Farm-to-Table Cooking Task: Environmental Variability and Strategic Flexibility.}
\noindent In Farm-to-Table Cooking task, agents must adapt their strategies to changing environmental conditions and varying difficulty levels. They need to gather information, source ingredients either from containers or through activities like harvesting and hunting, and adjust their methods to prepare complex dishes.

\noindent We simulate this by having agents act as farmers who are tasked with making \textbf{cake} and \textbf{rabbit stew} in Minecraft. These recipes are recognized for their high complexity in terms of ingredient synthesis, making them challenging targets for the task. Further details can be found in Appendix~\ref{Farm-to-Table Cooking illustrate}.

\paragraph{Escape Room Challenge Task: Synchronization and Sequential Execution.}
\noindent Escape Room Challenge task tests agents' ability to work together and perform actions in a precise order, focusing on synchronization and timing. Agents must navigate environments with objects that have specific activation requirements, and success depends on their coordinated timing and teamwork.

\noindent Each room offers unique challenges that demand effective team collaboration and strategic planning. For example, a basic task may require two agents to press switches at different locations simultaneously to open a door.
\noindent Further details and visual representations of each scenario can be found in Appendix \ref{Escape Room illustrate}.

\section{VillagerAgent: A Directed Acyclic Graph Multi-Agent Framework}
\subsection{Overview}

\noindent The VillagerAgent framework comprises four main components: Task Decomposer, Agent Controller, State Manager, and Base Agents. It operates by having the Task Decomposer generate a Directed Acyclic Graph (DAG) of subtask nodes each round, based on the current state, while the Agent Controller oversees the assignment of these subtasks to the Base Agents for execution and self-reflection. Meanwhile, the State Manager is responsible for maintaining the status information of both the environment and the agents (Figure \ref{fig:flowchart}).

\paragraph{Agent Notations.}
\noindent We denote each base agent as $A_i$ and the corresponding agent state as $S_i$. The agent state is a textual representation that recursively summarizes the agent's actions, possessions, and the entities in the surrounding environment.
Each agent has an action history (${H}_i$) that consists of the last $p$ actions.
We assume that there are $k$ agents in the game. The agent set can be represented as \(\mathbb{A}=\{A_i|i=1, \ldots, k\}\) and the agent state set is denoted as \(\mathbb{S} = \{S_i|i=1, \ldots, k\}\)

\paragraph{Task Notations.}
\noindent We model the execution dependencies of a complex task with a graph of subtasks.
Each subtask node \( N_j \) is represented by a quadruple, i.e,
(\(T_j, D_j, \mathbb{C}_j, F_j\)).
$T$ denotes the subtask description and $D$ represents the data from documents related to the subtask.
\(\mathbb{C}\) represents the assigned agents that have been selected by the Task Manager from the base agent set \(\mathbb{A}\).
$F$ denotes the execution feedback.
We denote the set of subtask nodes as  \(\mathbb{N}=\{N_j|j=1, \ldots, m\}\) where $m$ is the number of subtask nodes.

\subsection{Task Decomposer}

\noindent The Task Decomposer is responsible for managing and constructing the directed graph \( G \). 
The directed graph represents the concurrency of the subtasks. In this graph, each node \( v_i \in V \) corresponds to a subtask \( N_i \), and each directed edge \( (v_i, v_j)  \) signifies that subtask \( N_i \) must be completed before commencing subtask \( N_j \). Parallel execution of subtasks is permitted when there is no direct edge dictating the execution order between them.
The details of constructing the directed graph \(G\) from the set of subtasks \(\mathbb{N}\) can be found in Appendix~\ref{construct graph}.

\paragraph{Subtask Set Update.}
\noindent The Task Decomposer is also used to update the subtask set \(\mathbb{N}\). Given the goal task description $T_{g}$, the relevant environment state \(E\) queried from the State Manager, the agent state set \(\mathbb{S}\), and the current nodes \(\mathbb{N}\),
the Task Decomposer generates a set of new subtask nodes \(\mathbb{N'}\).

\[
\mathbb{N'} = \text{TD}(E, T_g, \mathbb{S}, \mathbb{N})
\]
\[
\mathbb{N} = \mathbb{N'} \cup  \mathbb{N}
\]
During task decomposition, the Task Decomposer adopts a zero-shot chain-of-thought (CoT) approach~\cite{wei2023chainofthought}. This method is integrated into the prompt, as Figure \ref{fig:Task Decomposer Prompt} illustrates, to guide the LLM in generating responses in JSON format, specify the index of the immediate predecessor for each subtask as needed and specify JSON path expressions for each subtask, referencing the provided data $D$. Subsequently, each subtask node will use these JSON path expressions to query the data related to its subtask.

\subsection{Agent Controller}

\noindent The Agent Controller focuses on analyzing the task graph and assigning the appropriate subtask to the right agent in an efficient manner.

\paragraph{Ready-to-Execute Tasks Identification.}
\noindent The Agent Controller identifies ready-to-execute task set \( \mathbb{N}_{ready}\).
It checks all unexecuted tasks, where tasks with no remaining dependencies will be added to the ready-to-execute task set  \( \mathbb{N}_{ready}\).

\paragraph{Subtask Allocation.}
\noindent Based on the environment state \(E\), ready-to-execute nodes \(\mathbb{N}_{ready}\), and the states of the agents \(\mathbb{S}\), the Agent Controller determines the allocation of agents to subtasks:
\[
\text{AC}(E, \mathbb{N}_{ready}, \mathbb{A}, \mathbb{S}) \rightarrow [(A_i, N_j), \ldots]
\]
In this process, the Agent Controller (AC) queries LLM to pair tasks with agents. It anticipates a JSON-formatted response containing the indices of tasks and the identifiers of the selected agents. The Agent Controller initiates the execution of tasks by the designated agents simultaneously.

\subsection{State Manager}
\noindent The State Manager (SM) is used to update the agent states and the environment information. 

\paragraph{Agent State Update.} 
\noindent SM updates the agent state based on the agent's action history \(H_i\):
\[ S_i = \text{LLM}(prompt_a, S_i, H_i). \]
where \(prompt_a\) is the agent state update prompt.
The agent state \(S_i\) acts as a long-term memory, in contrast to the action history \(H_i\), which serves as short-term memory.

\paragraph{Environment State Retrieval.}
\noindent The global environment state ($I$) is the union of the local environment state from each agent. The local environment state of agent $A_i$ can be obtained via the library API, i.e., \text{Env}(\(A_i\)).

\noindent Given the task description \( T_g \), the relevant environment state \( E \) can be retrieved from the global environment state ($I$):
\[ E = \text{LLM}(prompt_e, T_g, I). \]
where \(prompt_e\) is the environment state retrieval prompt.
\(prompt_a, prompt_e\) can be found in 
Appendix~\ref{fig:State Manager Agent State Update Prompt},~\ref{fig:Environment Summary Prompt}.

\subsection{Base Agent Architecture}
\noindent Each base agent \(A_i\) is responsible for executing its assigned subtask node \(N_j\). 
The states of the agents associated with the predecessor nodes of the current node $N_j$ in DAG can be represented as \( \mathbb{S}_{\text{selected}} \).
This execution results in an updated temporal action history and generates feedback:
\[
(H_i, F_j) = \text{Exec}(N_j, H_i, \mathbb{S}_{\text{selected}}, E)
\]
\noindent Upon execution of the subtask node \(N_j\), two processes occur within the agent \(A_i\):


\paragraph{ReAct Procedure.}
\noindent The Base Agent formulates a prompt that integrates its action history \(H_i\), the current state of agents \(\mathbb{S}_i\), the assigned subtask node \(N_j\), and environmental data \(E\) provided by the State Manager. Utilizing the ReAct method, the agent iteratively generates actions and observations.\cite{yao2023react} This iterative process is subject to a constraint of a maximum of 6 iterations or a total execution time limit of 120 seconds.

\paragraph{Self-Reflection.}
\noindent Upon completion of the task, the Base Agent updates the action history \(H_i\) and the task description \(T\) into a reflection prompt. LLM then generates a response that serves as feedback \(F_j\) for the subtask node \(N_j\).

\section{Experiments}

\begin{figure}[t]
  \centering 
  \includegraphics[width=\linewidth]{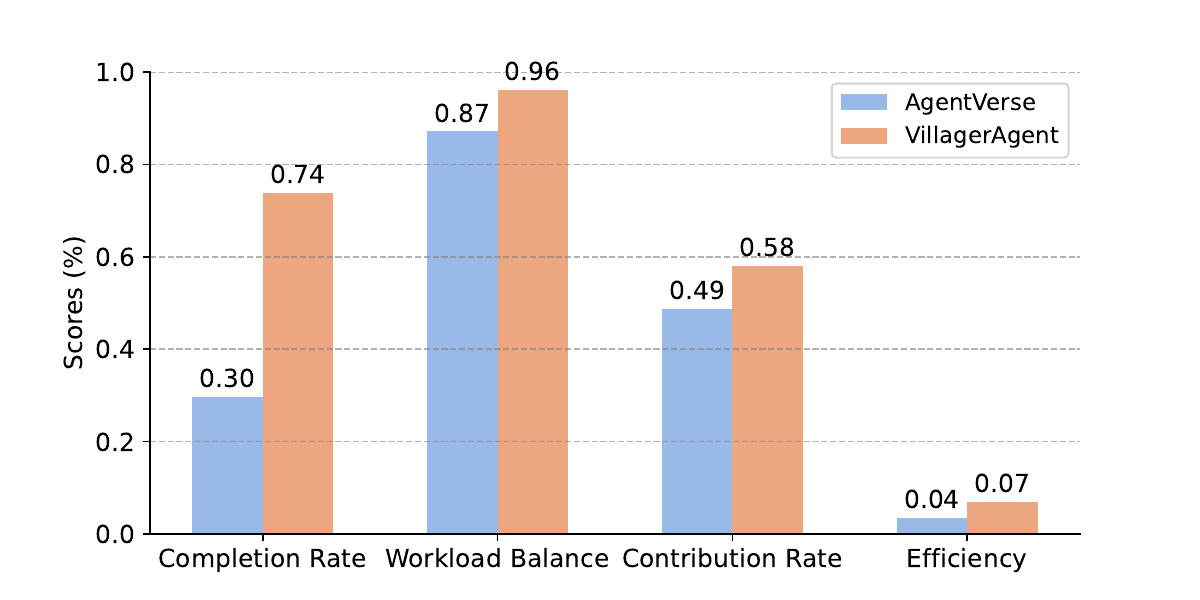} 
  \caption{Comparison of VillagerAgent and AgentVerse on Farm-to-Table Cooking Task. VillagerAgent outperforms AgentVerse in Completion Rate \citep{chen2023agentverse}. } 
  \label{fig:Comparison of VillagerAgent and AgentVerse Metrics} 
\end{figure}

\begin{figure}[t]
  \centering 
  \includegraphics[width=\linewidth]{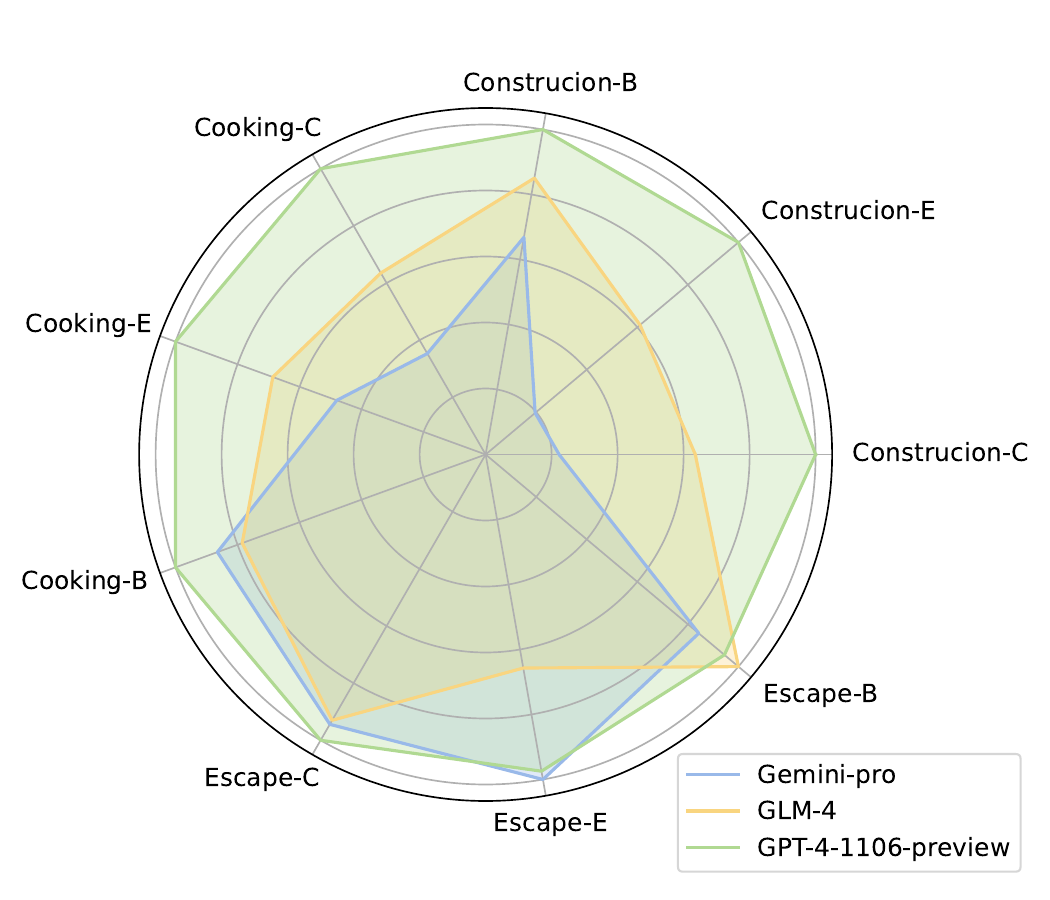} 
  \caption{Comparison of LLMs on VillagerBench. We show the relative performance gap
against the best in each scenario. GPT-4-1106-preview achieves higher scores across most metrics, whereas Gemini-Pro demonstrates better efficiency in the Escape Room Challenge.} 
  \label{fig:VillagerBench performance} 
\end{figure}

\paragraph{LLM Capability Test.}
\noindent To rigorously evaluate the capabilities of LLMs, we conducted tests on the VillagerBench benchmark using the VillagerAgent framework based on three models: GPT-4-1106-preview\cite{openai2023gpt4}, Gemini Pro\cite{geminiteam2023gemini}, and GLM-4\cite{du2022glm}. Our evaluation targeted three types of tasks: 100 Construction tasks, 100 Farm-to-table cooking tasks, and 25 Escape room challenges, each executed once. We terminate a testing round if the task execution exceeds the anticipated time frame or once the task has been successfully completed. The parameters for LLM reasoning can be found in Appendix~\ref{tab:model_config}.

\paragraph{Construction Cooperation Task.} 
\noindent For the construction tasks ranging from 0 to 99, we deployed two agents, Alice and Bob, each equipped with essential APIs, to collaborate effectively. We intentionally omitted the requirement for agents to mine blocks from the material factory, considering the inherent complexity of the tasks. The blueprint provided to the agents is a more concise and readable format, thereby streamlining the context and facilitating more efficient task completion, as detailed in Appendix~\ref{construction illustrate}.

\paragraph{Farm-to-Table Cooking Task.} 
\noindent For the Farm-to-Table Cooking tasks, numbered 0 through 99. Tasks 0 to 35 are dedicated to cake-making, while tasks 36 to 99 focus on the preparation of rabbit stew. We supply cooking recipes to serve as a reference for the agents. \textbf{VillagerAgent vs. AgentVerse in Cooking}: We've transitioned AgentVerse BaseAgent from the Voyager environment~\cite{wang2023voyager} to our VillagerBench BaseAgent, ensuring a fair comparison by preserving the prompt format and default settings, including the use of agent names Alice and Bob. Our modifications involve the adoption of the gpt-4-1106-preview language model, setting the temperature parameter to 0, and refining the feedback prompt to suit our ReAct Agent (Figure~\ref{fig:AgentVerse Config}).

\begin{table*}[h!]
\centering
\footnotesize
\begin{tabular}{@{}llllllll@{}}
\toprule
\multirow{2}{*}{\textbf{Models}} &  \multicolumn{4}{c}{\textbf{Construction Task Avg. Score}} & \multicolumn{3}{c}{\textbf{Escape Challenge Avg. Score}}\\ 
\cmidrule(lr){2-5} \cmidrule(lr){6-8}
& \textbf{C (\%)} & \textbf{VHR (\%)} &  \textbf{E (\%/min)}  &    \textbf{B (\%)} & \textbf{C (\%)} &  \textbf{E (\%/min)}  &  \textbf{B (\%)} \\
\midrule
gemini-pro & 8.12 & 13.83 & 0.76 & 63.74 & 69.2 & \textbf{153.3} & 80.35 \\ 
glm-4 & 23.16 & 29.36 & 2.37 & 81.12 & 68.17 & 100.8 & \textbf{95.3} \\ 
\textbf{gpt-4-1106-preview} & \textbf{36.45} & \textbf{49.05} & \textbf{3.88} & \textbf{95.38} & \textbf{73.29} & 149.4 & 90.03 \\ 
\textbf{gpt-4-1106-preview} (3-agents) & \textbf{52.17} & \textbf{61.02} & \textbf{6.26} & 89.83 & 69.78 & 227.4 & 67.01 \\ 
\bottomrule
\end{tabular}
\caption{GPT-4-1106-preview\cite{openai2023gpt4}, GLM-4\cite{du2022glm} and Gemini-Pro\cite{geminiteam2023gemini} results on Construction Cooperation task and Escape Room Challenge Task. (The Escape Room Challenge Task updates to accommodate varying numbers of agents.) }
\label{construction_escape}
\end{table*}

\paragraph{Escape Room Challenge Task.} 
\noindent We've crafted 18 atom-based escape room tasks that simulate puzzle-solving scenarios for agents. Our generator constructs these tasks from the ground up, selecting appropriate atom tasks based on room attributes, required materials, and agent information, and then automatically scales them into full-fledged puzzles. The generator also ensures task feasibility by accounting for agent cooperation and item dependencies. For consistent LLM testing, we've designated seeds for each of the five difficulty levels, with 25 unique tasks in total, and set a default simultaneous item activation wait time of 30 seconds for task completion.

\paragraph{Benchmarking Our VillagerAgent in Overcooked-AI.}
\noindent We conducted tests on VillagerAgent (equipped with GPT-4) within the Overcooked environment, following the methodology used in ProAgent\cite{zhang2024proagent}. We analyzed the prompt tokens for each test. Consistent with the settings outlined in ProAgent, we evaluated each layout across 5 episodes, with the horizon set to 400.

\paragraph{Influence of Agent Quantity on Cooperative Task Execution.}
\noindent We analyzed how varying numbers of agents (1, 2, 4, 8) affect cooperative task performance in construction scenarios, specifically comparing the simplest task(task 0) and a complex task(task 64). Using the GPT-4-1106-preview\cite{openai2023gpt4} model within the VillagerAgent framework, each task was repeated six times.

\paragraph{Assessing the Impact of Varied Agent Abilities on Cooperative Task Performance.}
\noindent We evaluate how different agent skill sets impact a complex farm-to-table cooking task (task 99 - rabbit stew preparation). With GPT-4-1106-preview\cite{openai2023gpt4} as the base model, we tested two trios of agents, each set consisted of three agents: one with uniform API abilities (7 Base APIs plus SmeltingCooking, MineBlock, and AttackTarget) and another with diverse abilities (7 Base APIs with one unique additional API per agent). Each repeated six times.

\subsection{Evaluation Metrics} 
\paragraph{Completion Rate (C).}
\noindent For each scenario, we monitor certain indicators that signify progress towards the scenario's objectives, such as blocks, ingredients or triggers. The completion rate is calculated based on the quantity of these indicators, providing a measure of how much of the scenario has been completed defined in Appendix\ref{sec:Metrics}. The formula for calculating the completion rate is as follows:
\[
\text{Completion (C)} = \frac{\text{\# Indicators Detected}}{\text{\# Total Indicators Expected}}
\]

\paragraph{Efficiency of Completion (E).}
\noindent It is defined as the ratio of the task completion rate to the actual time taken by the agents.
The efficiency of completion is computed as follows:
\[
\text{Efficiency (E)} = \frac{\text{\# Task Completion Rate}}{\text{\# Total Execution Time}}
\]


\paragraph{Balanced Agent Utilization Score (B).}
\noindent This metric assesses the distribution of workload among agents, aiming for a balanced utilization where each agent's active running time is similar. The ideal state is one where no single agent is either overburdened or underutilized.
\begin{equation}
\mathbf{t'} = \frac{\mathbf{t} - \min(\mathbf{t})}{\max(\mathbf{t}) - \min(\mathbf{t})}
\end{equation}
\begin{equation}
\text{Balance(B)} = 1 - \sigma(\mathbf{t'})
\end{equation}
Here, $n$ is the number of agents, $\mathbf{t}\in \mathbb{R}^n$, $\mathbf{t}_i$ represents the active running time of agent $i$, and $\bar{\mathbf{t}}$ is the average active running time across all agents.

\paragraph{Block Placement View Hit Rate (VHR).} 
\noindent evaluates the structural integrity and visual coherence of the construction from multiple vantage points. It is calculated as the intersection over union (IoU) of the constructed structure with the expected structure across a predefined set of viewpoints.
\begin{equation}
S_{vhr} = \frac{1}{V} \sum_{v=1}^{V} IoU(C_{v_{(\theta,\phi)}}, E_{v_{(\theta,\phi)}})
\end{equation}
Here, \( V \) is the number of viewpoints, \( C_v \) is the construction as seen from viewpoint \( v \), and \( E_v \) is the expected view from viewpoint \( v \).

\paragraph{Agent Contribution Rate (ACR).}
\noindent quantifies the contribution of each agent in a Minecraft game based on the items they have crafted in farm-to-table cooking tasks. 
The specific definitions can be found in Appendix~\ref{sec:Metrics}.

\begin{table}[t]
\small 
\setlength{\tabcolsep}{6pt} 
\begin{tabular}{@{}llllllll@{}}
\toprule
\multirow{2}{*}{\textbf{Agent Type}} &\multicolumn{4}{c}{\textbf{Farm-to-Table Cooking Avg. Score}}\\ 
\cmidrule(lr){2-5}
& \textbf{C (\%)} & \textbf{ACR(\%)} &  \textbf{E (\%/min)}  &    \textbf{B (\%)}\\
\midrule
  \textbf{Same} & \textbf{56.67} & \textbf{60.22} & \textbf{3.91} & \textbf{95.47} \\ 
Diverse & 36.67 & 30.46 & 2.87 & 92.2 \\ 

\bottomrule
\end{tabular}
\caption{Results of varied agent abilities
on cooperative task performance on Farm-to-Table Cooking Task 99.}

\label{agent ability}
\end{table}

\begin{table}[t]
\centering
\small 
\setlength{\tabcolsep}{6pt} 
\begin{tabular}{@{}p{2.2cm}llll@{}} 
\toprule
\multirow{2}{*}{\textbf{Models}} &  \multicolumn{4}{c}{\textbf{Cooking Task Avg. Score}}\\ 
\cmidrule(lr){2-5}
& \textbf{C (\%)} & \textbf{ACR} &  \textbf{E (\%/min)}  &    \textbf{B (\%)}\\
\midrule
AgentVerse gpt & 29.75 & 48.64 & 3.54 & 87.13 \\ 
VillagerAgent gemini & 26.05 & 32.92 & 3.35 & 83.15 \\ 
VillagerAgent glm & 46.84 & 54.07 & 4.79 & 75.46 \\ 
\textbf{VillagerAgent gpt (2-agents)} & 73.75 & \textbf{58.11} & 6.98 & \textbf{96.13} \\ 
\textbf{VillagerAgent gpt (3-agents)} & \textbf{85.26} & 55.60 & \textbf{21.90} & 84.38 \\ 
\bottomrule
\end{tabular}
\caption{Performance comparison between AgentVerse\cite{chen2023agentverse} and VillagerAgent on the Farm-to-Table Task. Note that gpt refers to GPT-4-1106-preview, gemini to Gemini-Pro, and glm to GLM-4}
\label{farming}
\end{table}

\subsection{Evaluation Results}


\paragraph{GPT-4 with VillagerAgent Achieves Optimal Performance.}
\noindent Across the board, GPT-4-1106-preview, when integrated with VillagerAgent, consistently delivered the highest completion scores in task allocation (Figure~\ref{fig:Comparison of VillagerAgent and AgentVerse Metrics}), as seen in Construction, Escape Room Tasks and Farm-to-Table Cooking (Table~\ref{construction_escape},~\ref{farming}). It demonstrated a superior understanding of task requirements and agent management, outperforming GLM-4 and Gemini-Pro in View Hit Rate (VHR) and Agent Contribution Rate (ACR).

\paragraph{Gemini-Pro Excels in Efficiency for Escape Room Challenge.}
\noindent In the context of less complex tasks that prioritize timing and sequence, such as the Escape Room Tasks, Gemini-Pro showcased its strengths. It achieved efficiency comparable to GLM-4 and, in some cases, outperformed others due to its faster inference and response times, leading to a high-efficiency rating (Table~\ref{construction_escape}).

\paragraph{VillagerAgent Outperforms AgentVerse.}
\noindent Despite both utilizing GPT-4, VillagerAgent outperforms AgentVerse in the Farm-to-Table Cooking Tasks(Figure~\ref{fig:Comparison of VillagerAgent and AgentVerse Metrics}), showing less hallucinatory behavior and a lower failure rate (18.2\% for VillagerAgent vs. 44.4\% for AgentVerse, as seen in Figure~\ref{fig:hallucination}). Although VillagerAgent uses more tokens on average (126 vs. 107 for AgentVerse), it achieves a significantly lower Token Cost (Avg. 1.79 vs. 10.3 for AgentVerse), indicating a more efficient use of resources for higher scores, as detailed in Table~\ref{task_difficulty}.

\paragraph{Agent Collaboration and Performance Dynamics.}
\noindent Data analysis from Table~\ref{agent num} shows that VillagerAgent's task performance improves with additional agents up to a point, after which it declines. Initially, more agents contribute positively, enhancing task handling through collective capability. However, as agent numbers increase further, performance gains diminish due to issues like resource competition and increased management complexity for the LLM. The relationship between agent count and performance is thus characterized by a peak at moderate levels of collaboration, suggesting an optimal balance for system efficiency without specifying a precise range.

\textbf{\begin{figure*}[h] 
  \centering 
  \includegraphics[width=\linewidth]{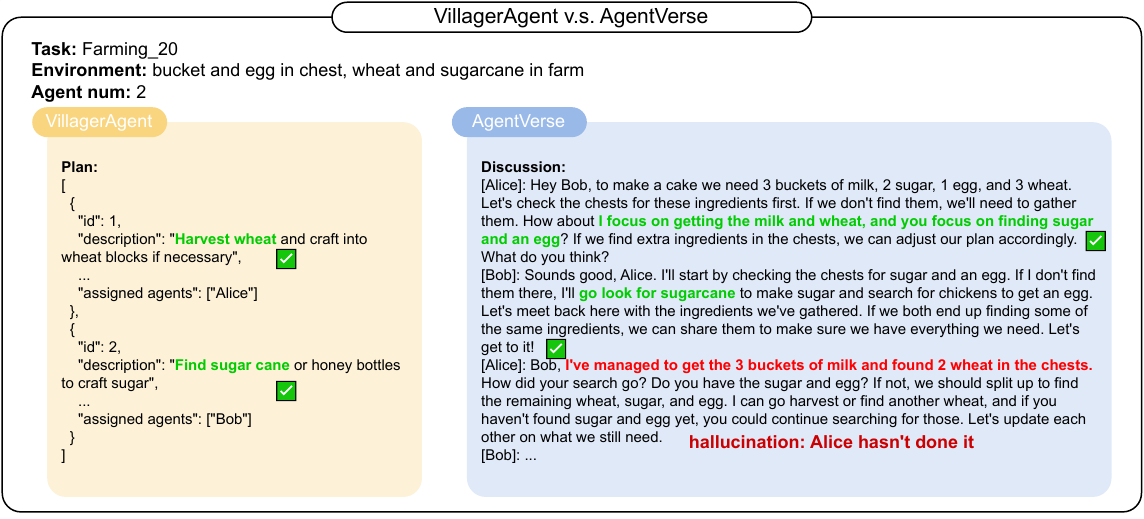} 
  \caption{VillagerAgent v.s. AgentVerse\cite{chen2023agentverse} on Farm-to-Table Task. Hallucination exists in agent discussion stage of AgentVerse.} 
  \label{fig:hallucination} 
\end{figure*}}

\paragraph{Diverse Abilities Hinder Coordination.}
\noindent The analysis of Table~\ref{agent ability} reveals that a trio of agents with distinct extra APIs underperforms in all evaluated metrics. This underperformance is attributed to the increased complexity in coordination when agents possess different capabilities. For example, the workflow may be disrupted if one agent's task depends on the completion of another's, leading to potential bottlenecks and task failure.

\noindent Despite the lower efficiency, the diverse skill set among agents introduces a richer complexity to the task environment, paving the way for more intricate cooperative interactions. While not optimal for score maximization, this setup serves as a fertile ground for investigating advanced collaborative behaviors and strategies within our benchmark framework.

\paragraph{Trade-off in Token Cost.}
\noindent To measure the relationship between task completion performance and token usage, we introduce the following formula to calculate the token cost.
\begin{equation}
Cost = \frac{CompletionTokens}{(Score + \varepsilon) + Action Num}
\end{equation}

\noindent Completion Tokens refers to the average token usage to complete each action.
Action Num refers to the number of valid actions executed during the task.
The score is the task score. We set epsilon=1 to prevent a score from dropping to zero.

\noindent We compute the Token Cost of VillagerAgent and AgentVerse on tasks of varying difficulty.
\begin{table}[ht]
\centering
\begin{tabular}{@{}llll@{}}
\toprule
Difficulty & Framework  & Tokens $\downarrow$ & Cost $\downarrow$ \\ \midrule
Easy            & AgentVerse & 109.26            & 17.76                   \\
Medium          & AgentVerse & 108.95            & 6.05                    \\
Hard            & AgentVerse & 101.33            & 4.32                    \\
Average         & AgentVerse & 107.17           & 10.30                   \\ \midrule
Easy            & VillagerAgent     & 122.33            & 1.73                    \\
Medium          & VillagerAgent     & 126.52            & 1.69                    \\
Hard            & VillagerAgent     & 131.13            & 2.01                    \\
Average         & VillagerAgent     & 126.00            & 1.79                    \\ \bottomrule
\end{tabular}
\caption{Comparison of Trade-off in Token Cost}
\label{task_difficulty}
\end{table}

\noindent We discovered that, although we use slightly more tokens per action compared to AgentVerse, our Token Cost is significantly lower. This indicates that the benefits we gain in terms of score outweigh the additional tokens we utilize.

\subsection{In Overcooked-AI}

\paragraph{VillagerAgent Outperforms ProAgent.}

\begin{figure*}[h]
  \centering 
  \includegraphics[width=\linewidth ]{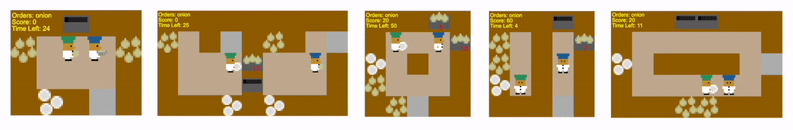} 
  \caption{Overcooked-AI Scenarios} 
  \label{overcooked_ai} 
\end{figure*}

\noindent In Overcooked-AI~\ref{overcooked_ai}, our VillagerAgent (w/gpt-4), surpasses ProAgent\cite{zhang2024proagent} across all five scenarios: Cramped Room, Asymmetric Advantages, Coordination Ring, Forced Coordination, and Counter Circuit. Each scenario tests different aspects of cooperative strategy and efficiency in a shared task environment. Notably excelling in the Forced Coordination scenario—a highly interdependent task requiring material sharing in confined spaces. This superior performance is attributed to our use of directed acyclic graphs for task management, enhancing efficiency in complex cooperative tasks, as detailed in Table~\ref{performance_comparison}.


\paragraph{Efficiency and Transferability of Prompts.}
\noindent we compare ProAgent and VillagerAgent regarding the use of prompt. The results are shown in the table below. Our framework utilized a single set of prompts to accomplish five tasks in Overcooked, whereas ProAgent employed five specific sets of prompts. Similarly, we also used a single set of prompts across three scenarios in VillagerBench. We also observe that VillagerAgent uses fewer tokens in each test, implying its lower overhead and better transferability.

\begin{table*}[ht]
\small
\centering

\begin{tabular}{@{}lcccccc@{}}
\toprule
Layout                & PBT           & FCP           & MEP           & COLE          & ProAgent      & VillagerAgent (ours) \\ \midrule
Cramped Room          & 178.8 ± 16.5  & 196.3 ± 16.8  & 185 ± 15      & 163.8 ± 24.1  & 197.3 ± 6.1   & \textbf{213.3 ± 9.43}  \\
Asymmetric Advantages & 182.2 ± 27.9  & 185.7 ± 22.7  & 155.7 ± 63.9  & 201.3 ± 34.5  & 228.7 ± 23    & \textbf{304 ± 8.76}    \\
Coordination Ring     & 141.3 ± 28    & 148.8 ± 19.4  & 167.2 ± 22.4  & 168.8 ± 26.1  & 175.3 ± 29    & \textbf{226.7 ± 18.9} \\
Forced Coordination   & 15.3 ± 17.1   & 44.7 ± 36.4   & 23.3 ± 19.8   & 24 ± 21.8     & 49.7 ± 33.1   & \textbf{120 ± 16.97}   \\
Counter Circuit       & 64.7 ± 45.9   & 58.3 ± 37.5   & 74.3 ± 39.1   & 95.5 ± 25.2   & 126.3 ± 32.3  & \textbf{148 ± 4.38}    \\ \bottomrule
\end{tabular}
\caption{Performance comparison of VillagerAgent and ProAgent across different scenarios in Overcooked-AI.}
\label{performance_comparison}

\end{table*}

\section{Related Work}

\paragraph{Minecraft Agents.}
\noindent Minecraft agents are intelligent programs that can perform various tasks within the Minecraft world. Recently, researchers have come to be aware of the extraordinary general planning ability of LLMs \cite{huang2022language}.
Many works \cite{huang2022inner,yuan2023skill,wang2023describe,wang2023voyager,zhu2023ghost} have leveraged LLMs for enhancing the high-level planning ability of
Minecraft agents. Inner Monologue \cite{huang2022inner} leveraged environment feedback to improve the planning
ability of LLM. Voyager \cite{wang2023voyager} developed a skill library of executable code for storing and retrieving behaviors. 
The base agent in our VillagerAgent framework accounts for the states of other agents and features a modular design, enabling it to function independently or in collaboration with other agents.

\paragraph{MultiAgent Frameworks.}
\noindent MultiAgent frameworks are increasingly leveraging LLMs due to their potential in complex system development \cite{qian2023experiential,qian2023communicative,xie2023openagents,wu2023autogen}. CAMEL utilizes role-play to reduce hallucinations and improve collaboration \cite{li2023camel}. MindAgent's CuisineWorld uses a Collaboration Score to gauge team efficiency \cite{gong2023mindagent}. DEPS further extended this closed-loop interaction by introducing description, explainer and selector \cite{wang2023describe}. AgentVerse structures its system into recruitment, decision-making, execution, and evaluation, optimizing group performance \cite{chen2023agentverse}. MetaGPT adopts an assembly line method, assigning roles to streamline task completion \cite{hong2023metagpt}. However, these frameworks often face limitations in task flexibility and scalability\cite{gong2023mindagent,chen2023agentverse,hong2023metagpt}. Our VillagerAgent framework improves collaborative efficiency for complex tasks by modeling task graphs.

\paragraph{LLM-as-Agent Benchmarks.}
\noindent Recent studies highlight the potential of Large Language Models (LLMs) as agents capable of tool use \cite{wang2023survey,xi2023rise}. Emerging benchmarks aim to rigorously evaluate these models' performance \cite{liu2023agentbench,xu2023magic,carroll2020utility,huang2023benchmarking,wu2023smartplay,ruan2023identifying}. MCU\cite{lin2023mcu} discusses a method that uses atomic tasks as basic components to create a wide range of tasks (SkillForge). Our research focuses on multi-agent, enhances task complexity using Minecraft commands, introduces more intricate challenges like long-distance switch activation.
The Overcooked environment is notable for coordination experiments \cite{carroll2020utility}, while MAgIC focuses on assessing LLMs' cognitive and collaborative abilities in text-based multi-agent settings \cite{xu2023magic}. Existing benchmarks, however, may not fully capture the capabilities of LLMs as multi-agents. Inspired by multiple single-agent studies conducted within Minecraft.\cite{huang2022inner,yuan2023skill,wang2023describe,wang2023voyager,zhu2023ghost}
Our VillagerBench leverages Minecraft's API to create domains that mimic real-world tasks, facilitating multi-agent system evaluation and research advancement.


\section{Conclusion}
\noindent In this study, we introduce VillagerBench, a Minecraft multi-agent benchmark platform. We design three distinct scenarios within VillagerBench to evaluate collaborative tasks, aiming to assess the performance of our VillagerAgent framework. Our framework employs Directed Acyclic Graphs (DAG) to decompose tasks, enabling efficient and coordinated execution by agents. We benchmark the coordination skills of three LLMs using these metrics and demonstrate the effectiveness of our VillagerAgent framework. We also explore how agent count and capability diversity impact framework performance.

\section*{Limitations}

\noindent Our VillagerAgent framework, while improving performance within the Minecraft multi-agent benchmark (VillagerBench), encounters a low overall task completion rate. This is partly due to the inherent complexity of the benchmark, which necessitates the use of a wide array of APIs, thereby enlarging the exploration space and complicating the execution of tasks, especially when agents have varied abilities.

\noindent One of the primary challenges is managing agents with varying capabilities, as it necessitates advanced coordination and balancing strategies to ensure effective teamwork. Our framework's performance diminishes when scaling beyond eight agents, suggesting issues with resource allocation and inter-agent communication efficiency. This decline could be attributed to the increased context length and the complexity of generating task graphs for a larger number of agents, analogous to a leader struggling to manage an excessive number of workers.

\noindent \textbf{Challenges in Practice.} If our work were to be applied in real-world settings, we anticipate facing challenges such as the complexity of processing dynamic information, variability in agent failures, and issues with the interpretability of Language Learning Models' outputs. Addressing these challenges may become one of the key directions for further research.

\section*{Acknowledgements}

This work is supported by the National Science and Technology Major Project (2022ZD0117802). This work is also supported by the General Program of the National Natural Science Foundation of China (62372403).

\bibliography{anthology,custom}

\begin{thebibliography}{30}
\expandafter\ifx\csname natexlab\endcsname\relax\def\natexlab#1{#1}\fi

\bibitem[{gem(2023)}]{geminiteam2023gemini}
 2023.
\newblock \href {http://arxiv.org/abs/2312.11805} {Gemini: A family of highly
  capable multimodal models}.

\bibitem[{ope(2023)}]{openai2023gpt4}
 2023.
\newblock \href {http://arxiv.org/abs/2303.08774} {Gpt-4 technical report}.

\bibitem[{Carroll et~al.(2020)Carroll, Shah, Ho, Griffiths, Seshia, Abbeel, and
  Dragan}]{carroll2020utility}
Micah Carroll, Rohin Shah, Mark~K. Ho, Thomas~L. Griffiths, Sanjit~A. Seshia,
  Pieter Abbeel, and Anca Dragan. 2020.
\newblock \href {http://arxiv.org/abs/1910.05789} {On the utility of learning
  about humans for human-ai coordination}.

\bibitem[{Chen et~al.(2023)Chen, Su, Zuo, Yang, Yuan, Qian, Chan, Qin, Lu, Xie
  et~al.}]{chen2023agentverse}
Weize Chen, Yusheng Su, Jingwei Zuo, Cheng Yang, Chenfei Yuan, Chen Qian,
  Chi-Min Chan, Yujia Qin, Yaxi Lu, Ruobing Xie, et~al. 2023.
\newblock Agentverse: Facilitating multi-agent collaboration and exploring
  emergent behaviors in agents.
\newblock \emph{arXiv preprint arXiv:2308.10848}.

\bibitem[{Du et~al.(2022)Du, Qian, Liu, Ding, Qiu, Yang, and Tang}]{du2022glm}
Zhengxiao Du, Yujie Qian, Xiao Liu, Ming Ding, Jiezhong Qiu, Zhilin Yang, and
  Jie Tang. 2022.
\newblock \href {http://arxiv.org/abs/2103.10360} {Glm: General language model
  pretraining with autoregressive blank infilling}.

\bibitem[{Gong et~al.(2023)Gong, Huang, Ma, Vo, Durante, Noda, Zheng, Zhu,
  Terzopoulos, Fei-Fei, and Gao}]{gong2023mindagent}
Ran Gong, Qiuyuan Huang, Xiaojian Ma, Hoi Vo, Zane Durante, Yusuke Noda, Zilong
  Zheng, Song-Chun Zhu, Demetri Terzopoulos, Li~Fei-Fei, and Jianfeng Gao.
  2023.
\newblock \href {http://arxiv.org/abs/2309.09971} {Mindagent: Emergent gaming
  interaction}.

\bibitem[{Hong et~al.(2023)Hong, Zhuge, Chen, Zheng, Cheng, Zhang, Wang, Wang,
  Yau, Lin, Zhou, Ran, Xiao, Wu, and Schmidhuber}]{hong2023metagpt}
Sirui Hong, Mingchen Zhuge, Jonathan Chen, Xiawu Zheng, Yuheng Cheng, Ceyao
  Zhang, Jinlin Wang, Zili Wang, Steven Ka~Shing Yau, Zijuan Lin, Liyang Zhou,
  Chenyu Ran, Lingfeng Xiao, Chenglin Wu, and Jürgen Schmidhuber. 2023.
\newblock \href {http://arxiv.org/abs/2308.00352} {Metagpt: Meta programming
  for a multi-agent collaborative framework}.

\bibitem[{Huang et~al.(2023)Huang, Vora, Liang, and
  Leskovec}]{huang2023benchmarking}
Qian Huang, Jian Vora, Percy Liang, and Jure Leskovec. 2023.
\newblock \href {http://arxiv.org/abs/2310.03302} {Benchmarking large language
  models as ai research agents}.

\bibitem[{Huang et~al.(2022{\natexlab{a}})Huang, Abbeel, Pathak, and
  Mordatch}]{huang2022language}
Wenlong Huang, Pieter Abbeel, Deepak Pathak, and Igor Mordatch.
  2022{\natexlab{a}}.
\newblock \href {http://arxiv.org/abs/2201.07207} {Language models as zero-shot
  planners: Extracting actionable knowledge for embodied agents}.

\bibitem[{Huang et~al.(2022{\natexlab{b}})Huang, Xia, Xiao, Chan, Liang,
  Florence, Zeng, Tompson, Mordatch, Chebotar, Sermanet, Brown, Jackson, Luu,
  Levine, Hausman, and Ichter}]{huang2022inner}
Wenlong Huang, Fei Xia, Ted Xiao, Harris Chan, Jacky Liang, Pete Florence, Andy
  Zeng, Jonathan Tompson, Igor Mordatch, Yevgen Chebotar, Pierre Sermanet, Noah
  Brown, Tomas Jackson, Linda Luu, Sergey Levine, Karol Hausman, and Brian
  Ichter. 2022{\natexlab{b}}.
\newblock \href {http://arxiv.org/abs/2207.05608} {Inner monologue: Embodied
  reasoning through planning with language models}.

\bibitem[{Li et~al.(2023)Li, Hammoud, Itani, Khizbullin, and
  Ghanem}]{li2023camel}
Guohao Li, Hasan Abed Al~Kader Hammoud, Hani Itani, Dmitrii Khizbullin, and
  Bernard Ghanem. 2023.
\newblock Camel: Communicative agents for "mind" exploration of large language
  model society.
\newblock In \emph{Thirty-seventh Conference on Neural Information Processing
  Systems}.

\bibitem[{Lin et~al.(2023)Lin, Wang, Ma, and Liang}]{lin2023mcu}
Haowei Lin, Zihao Wang, Jianzhu Ma, and Yitao Liang. 2023.
\newblock \href {http://arxiv.org/abs/2310.08367} {Mcu: A task-centric
  framework for open-ended agent evaluation in minecraft}.

\bibitem[{Liu et~al.(2023)Liu, Yu, Zhang, Xu, Lei, Lai, Gu, Ding, Men, Yang,
  Zhang, Deng, Zeng, Du, Zhang, Shen, Zhang, Su, Sun, Huang, Dong, and
  Tang}]{liu2023agentbench}
Xiao Liu, Hao Yu, Hanchen Zhang, Yifan Xu, Xuanyu Lei, Hanyu Lai, Yu~Gu,
  Hangliang Ding, Kaiwen Men, Kejuan Yang, Shudan Zhang, Xiang Deng, Aohan
  Zeng, Zhengxiao Du, Chenhui Zhang, Sheng Shen, Tianjun Zhang, Yu~Su, Huan
  Sun, Minlie Huang, Yuxiao Dong, and Jie Tang. 2023.
\newblock Agentbench: Evaluating llms as agents.
\newblock \emph{arXiv preprint arXiv: 2308.03688}.

\bibitem[{PrismarineJS(2013)}]{mineflayer}
PrismarineJS. 2013.
\newblock \href {https://github.com/PrismarineJS/mineflayer/tree/master}
  {Prismarinejs/mineflayer: Create minecraft bots with a powerful, stable, and
  high level javascript api.}

\bibitem[{Qian et~al.(2023{\natexlab{a}})Qian, Cong, Liu, Yang, Chen, Su, Dang,
  Li, Xu, Li, Liu, and Sun}]{qian2023communicative}
Chen Qian, Xin Cong, Wei Liu, Cheng Yang, Weize Chen, Yusheng Su, Yufan Dang,
  Jiahao Li, Juyuan Xu, Dahai Li, Zhiyuan Liu, and Maosong Sun.
  2023{\natexlab{a}}.
\newblock \href {http://arxiv.org/abs/2307.07924} {Communicative agents for
  software development}.

\bibitem[{Qian et~al.(2023{\natexlab{b}})Qian, Dang, Li, Liu, Chen, Yang, Liu,
  and Sun}]{qian2023experiential}
Chen Qian, Yufan Dang, Jiahao Li, Wei Liu, Weize Chen, Cheng Yang, Zhiyuan Liu,
  and Maosong Sun. 2023{\natexlab{b}}.
\newblock \href {http://arxiv.org/abs/2312.17025} {Experiential co-learning of
  software-developing agents}.

\bibitem[{Ruan et~al.(2023)Ruan, Dong, Wang, Pitis, Zhou, Ba, Dubois, Maddison,
  and Hashimoto}]{ruan2023identifying}
Yangjun Ruan, Honghua Dong, Andrew Wang, Silviu Pitis, Yongchao Zhou, Jimmy Ba,
  Yann Dubois, Chris~J. Maddison, and Tatsunori Hashimoto. 2023.
\newblock \href {http://arxiv.org/abs/2309.15817} {Identifying the risks of lm
  agents with an lm-emulated sandbox}.

\bibitem[{Wang et~al.(2023{\natexlab{a}})Wang, Xie, Jiang, Mandlekar, Xiao,
  Zhu, Fan, and Anandkumar}]{wang2023voyager}
Guanzhi Wang, Yuqi Xie, Yunfan Jiang, Ajay Mandlekar, Chaowei Xiao, Yuke Zhu,
  Linxi Fan, and Anima Anandkumar. 2023{\natexlab{a}}.
\newblock \href {http://arxiv.org/abs/2305.16291} {Voyager: An open-ended
  embodied agent with large language models}.

\bibitem[{Wang et~al.(2023{\natexlab{b}})Wang, Ma, Feng, Zhang, Yang, Zhang,
  Chen, Tang, Chen, Lin, Zhao, Wei, and Wen}]{wang2023survey}
Lei Wang, Chen Ma, Xueyang Feng, Zeyu Zhang, Hao Yang, Jingsen Zhang, Zhiyuan
  Chen, Jiakai Tang, Xu~Chen, Yankai Lin, Wayne~Xin Zhao, Zhewei Wei, and
  Ji-Rong Wen. 2023{\natexlab{b}}.
\newblock \href {http://arxiv.org/abs/2308.11432} {A survey on large language
  model based autonomous agents}.

\bibitem[{Wang et~al.(2023{\natexlab{c}})Wang, Cai, Chen, Liu, Ma, and
  Liang}]{wang2023describe}
Zihao Wang, Shaofei Cai, Guanzhou Chen, Anji Liu, Xiaojian Ma, and Yitao Liang.
  2023{\natexlab{c}}.
\newblock \href {http://arxiv.org/abs/2302.01560} {Describe, explain, plan and
  select: Interactive planning with large language models enables open-world
  multi-task agents}.

\bibitem[{Wei et~al.(2023)Wei, Wang, Schuurmans, Bosma, Ichter, Xia, Chi, Le,
  and Zhou}]{wei2023chainofthought}
Jason Wei, Xuezhi Wang, Dale Schuurmans, Maarten Bosma, Brian Ichter, Fei Xia,
  Ed~Chi, Quoc Le, and Denny Zhou. 2023.
\newblock \href {http://arxiv.org/abs/2201.11903} {Chain-of-thought prompting
  elicits reasoning in large language models}.

\bibitem[{Wu et~al.(2023{\natexlab{a}})Wu, Bansal, Zhang, Wu, Li, Zhu, Jiang,
  Zhang, Zhang, Liu, Awadallah, White, Burger, and Wang}]{wu2023autogen}
Qingyun Wu, Gagan Bansal, Jieyu Zhang, Yiran Wu, Beibin Li, Erkang Zhu,
  Li~Jiang, Xiaoyun Zhang, Shaokun Zhang, Jiale Liu, Ahmed~Hassan Awadallah,
  Ryen~W White, Doug Burger, and Chi Wang. 2023{\natexlab{a}}.
\newblock \href {http://arxiv.org/abs/2308.08155} {Autogen: Enabling next-gen
  llm applications via multi-agent conversation}.

\bibitem[{Wu et~al.(2023{\natexlab{b}})Wu, Tang, Mitchell, and
  Li}]{wu2023smartplay}
Yue Wu, Xuan Tang, Tom~M. Mitchell, and Yuanzhi Li. 2023{\natexlab{b}}.
\newblock \href {http://arxiv.org/abs/2310.01557} {Smartplay: A benchmark for
  llms as intelligent agents}.

\bibitem[{Xi et~al.(2023)Xi, Chen, Guo, He, Ding, Hong, Zhang, Wang, Jin, Zhou,
  Zheng, Fan, Wang, Xiong, Zhou, Wang, Jiang, Zou, Liu, Yin, Dou, Weng, Cheng,
  Zhang, Qin, Zheng, Qiu, Huang, and Gui}]{xi2023rise}
Zhiheng Xi, Wenxiang Chen, Xin Guo, Wei He, Yiwen Ding, Boyang Hong, Ming
  Zhang, Junzhe Wang, Senjie Jin, Enyu Zhou, Rui Zheng, Xiaoran Fan, Xiao Wang,
  Limao Xiong, Yuhao Zhou, Weiran Wang, Changhao Jiang, Yicheng Zou, Xiangyang
  Liu, Zhangyue Yin, Shihan Dou, Rongxiang Weng, Wensen Cheng, Qi~Zhang,
  Wenjuan Qin, Yongyan Zheng, Xipeng Qiu, Xuanjing Huang, and Tao Gui. 2023.
\newblock \href {http://arxiv.org/abs/2309.07864} {The rise and potential of
  large language model based agents: A survey}.

\bibitem[{Xie et~al.(2023)Xie, Zhou, Cheng, Shi, Weng, Liu, Hua, Zhao, Liu,
  Liu, Liu, Xu, Su, Shin, Xiong, and Yu}]{xie2023openagents}
Tianbao Xie, Fan Zhou, Zhoujun Cheng, Peng Shi, Luoxuan Weng, Yitao Liu,
  Toh~Jing Hua, Junning Zhao, Qian Liu, Che Liu, Leo~Z. Liu, Yiheng Xu, Hongjin
  Su, Dongchan Shin, Caiming Xiong, and Tao Yu. 2023.
\newblock \href {http://arxiv.org/abs/2310.10634} {Openagents: An open platform
  for language agents in the wild}.

\bibitem[{Xu et~al.(2023)Xu, Hu, Zhou, Ren, Dong, Keutzer, Ng, and
  Feng}]{xu2023magic}
Lin Xu, Zhiyuan Hu, Daquan Zhou, Hongyu Ren, Zhen Dong, Kurt Keutzer, See~Kiong
  Ng, and Jiashi Feng. 2023.
\newblock Magic: Benchmarking large language model powered multi-agent in
  cognition, adaptability, rationality and collaboration.
\newblock \emph{arXiv preprint arXiv: 2311.08562}.

\bibitem[{Yao et~al.(2023)Yao, Zhao, Yu, Du, Shafran, Narasimhan, and
  Cao}]{yao2023react}
Shunyu Yao, Jeffrey Zhao, Dian Yu, Nan Du, Izhak Shafran, Karthik Narasimhan,
  and Yuan Cao. 2023.
\newblock \href {http://arxiv.org/abs/2210.03629} {React: Synergizing reasoning
  and acting in language models}.

\bibitem[{Yuan et~al.(2023)Yuan, Zhang, Wang, Xie, Cai, Dong, and
  Lu}]{yuan2023skill}
Haoqi Yuan, Chi Zhang, Hongcheng Wang, Feiyang Xie, Penglin Cai, Hao Dong, and
  Zongqing Lu. 2023.
\newblock \href {http://arxiv.org/abs/2303.16563} {Skill reinforcement learning
  and planning for open-world long-horizon tasks}.

\bibitem[{Zhang et~al.(2024)Zhang, Yang, Hu, Wang, Li, Sun, Zhang, Zhang, Liu,
  Zhu, Chang, Zhang, Yin, Liang, and Yang}]{zhang2024proagent}
Ceyao Zhang, Kaijie Yang, Siyi Hu, Zihao Wang, Guanghe Li, Yihang Sun, Cheng
  Zhang, Zhaowei Zhang, Anji Liu, Song-Chun Zhu, Xiaojun Chang, Junge Zhang,
  Feng Yin, Yitao Liang, and Yaodong Yang. 2024.
\newblock \href {http://arxiv.org/abs/2308.11339} {Proagent: Building proactive
  cooperative agents with large language models}.

\bibitem[{Zhu et~al.(2023)Zhu, Chen, Tian, Tao, Su, Yang, Huang, Li, Lu, Wang,
  Qiao, Zhang, and Dai}]{zhu2023ghost}
Xizhou Zhu, Yuntao Chen, Hao Tian, Chenxin Tao, Weijie Su, Chenyu Yang, Gao
  Huang, Bin Li, Lewei Lu, Xiaogang Wang, Yu~Qiao, Zhaoxiang Zhang, and Jifeng
  Dai. 2023.
\newblock \href {http://arxiv.org/abs/2305.17144} {Ghost in the minecraft:
  Generally capable agents for open-world environments via large language
  models with text-based knowledge and memory}.

\end{thebibliography}
\bibliographystyle{acl_natbib}

\appendix

\newpage
\section{Metrics}
\label{sec:Metrics}

\subsection{Task Node Graph relevant algorithm}

\paragraph{Convert subtask node set to Graph.} 
\noindent Since LLMs are autoregressive, their outputs for subtasks often exhibit causal relationships. Leveraging this, we can assume that a given prompt suggests subsequent subtasks depend on or run concurrently with earlier ones, forming the basis for transforming them into a graph.

\noindent Task Decomposer construct graph using algorithm~\ref{algorithm1} to connect nodes representing subtasks:
\label{construct graph}
\begin{enumerate}
    \item Initialize the graph \(G\) with an empty set of vertices \(V\), an empty set of edges \(E\) and the input list of subtask nodes \(L\) containing \(N_1, N_2, \ldots, N_n\).
    \item Iterate over each node \(N_i\) in the list \(L\), where \(i\) ranges from 1 to \(n\). Then add the current node \(N_i\) to the vertex set \(V\).
    \item Check if the current node \(N_i\) has predecessor nodes \(P(N_i)\):
    \begin{itemize}
        \item If \(N_i\) has predecessors, for each predecessor node \(p_j\), add an edge from \(p_j\) to \(N_i\) to the edge set \(E\).
        \item If \(N_i\) does not have predecessors and \(i > 1\), implying it may share predecessors with the previous node \(N_{i-1}\), for each predecessor of \(N_{i-1}\), \(p_k\), add an edge from \(p_k\) to \(N_i\) to the edge set \(E\).
    \end{itemize}
    \item Repeat steps 2 and 3 until all nodes in the list have been processed.
\end{enumerate}

\begin{figure*}[tp]
  \centering 
  \includegraphics[width=\linewidth]{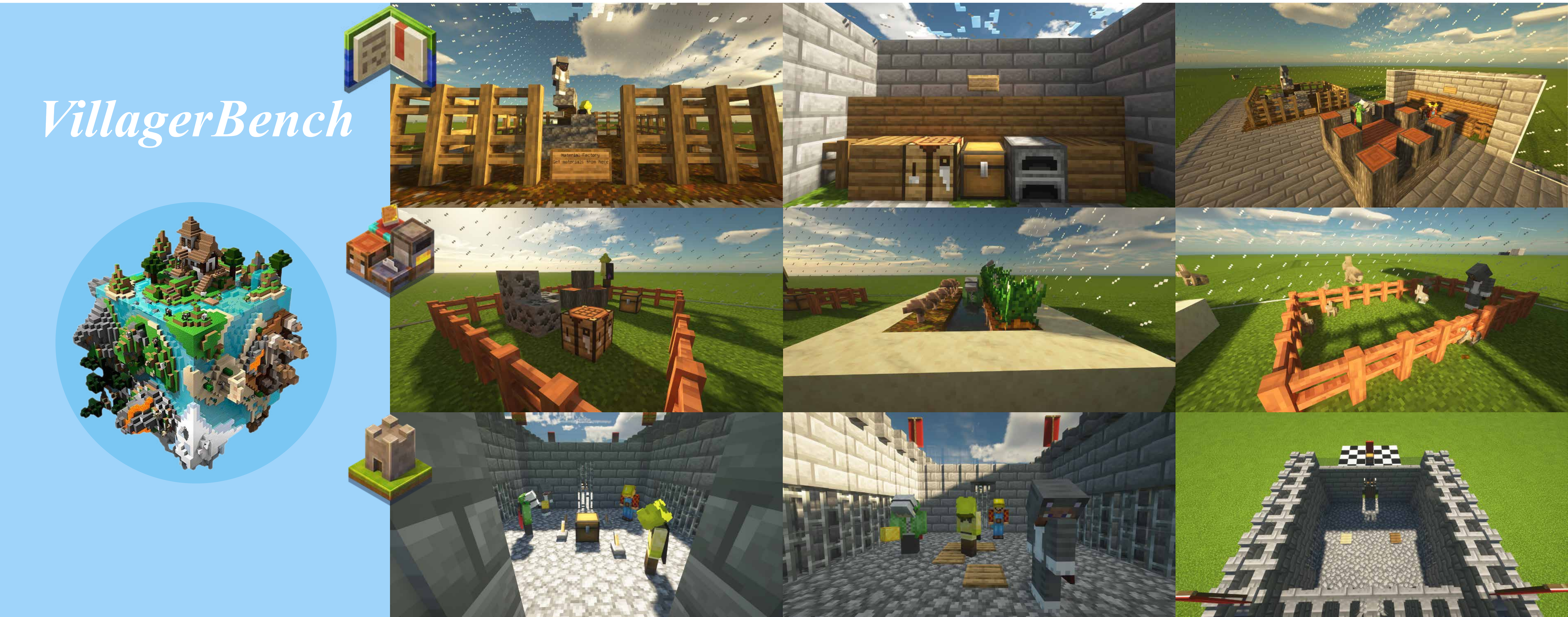} 
  \caption{Live demonstration of agents performing tasks in VillagerBench scenarios.} 
  \label{fig:display} 
\end{figure*}

\begin{algorithm}
\caption{Convert Task List to Graph}
\label{algorithm1}
\begin{algorithmic}[1]
\State $G \gets (V, E)$ with $V \gets \emptyset$, $E \gets \emptyset$
\State $L \gets [N_1, N_2, \ldots, N_n]$ \Comment{Input list}
\For{$i \gets 1$ to $n$}
    \State $V \gets V \cup \{N_i\}$ \Comment{Add element as a node}
    \If{$P(N_i) \neq \emptyset$}
        \ForAll{$p_j \in P(N_i)$}
            \State $E \gets E \cup \{(p_j, N_i)\}$ \Comment{Add edges from predecessors}
        \EndFor
    \ElsIf{$i > 1$}
        \ForAll{$p_k \in P(N_{i-1})$}
            \State $E \gets E \cup \{(p_k, N_i)\}$ \Comment{Share predecessors with previous element}
        \EndFor
    \EndIf
\EndFor
\end{algorithmic}
\end{algorithm}

\begin{algorithm}
\caption{Find Ready-to-Execute Tasks}
\label{algorithm2}
\begin{algorithmic}[1]
\Require $G = (V, E)$ \Comment{Task graph with nodes and edges}
\Require $S \subseteq V$ \Comment{Set of successfully executed tasks}
\Require $U \subseteq V$ \Comment{Set of unexecuted tasks}
\State $R \gets \emptyset$ \Comment{Result set of ready-to-execute tasks}
\ForAll{$N_i \in U$}
    \State $P(N_i) \gets \{p_j \mid (p_j, N_i) \in E\}$ \Comment{Find predecessors of $N_i$}
    \If{$P(N_i) = \emptyset$ or $P(N_i) \subseteq S$}
        \State $R \gets R \cup \{N_i\}$ \Comment{Add if no predecessors or all predecessors executed}
    \EndIf
\EndFor
\State \Return $R$
\end{algorithmic}
\end{algorithm}

\subsection{Construction Task Complete Rate (C)}
\paragraph{Construction Task Complete Rate.}
\noindent quantifies the alignment of the constructed structure with the provided blueprint. It is defined as the ratio of correctly placed blocks to the total number of blocks specified by the blueprint. A higher $C$ indicates a closer match to the intended design, reflecting the agents' ability to accurately interpret and execute the construction plan.
\begin{equation}
C = \frac{|P_{(x,y,z,\theta,\phi)} \cap B_{(x,y,z,\theta,\phi)}|}{|B_{(x,y,z,\theta,\phi)}|}
\end{equation}
Here, \( P \) represents the set of placed blocks, and \( B \) represents the set of blocks in the blueprint. $\theta$ denotes facing and $\phi$ denotes axis.


\subsection{Construction Dependency Complexity (D)}
\noindent 
\begin{equation}
 D = \sum_{i=1}^{B} \left( \frac{1}{EP_i} + W_h (H_i - G) \right) + D_i
\end{equation}
Here, \(EP\) represents the effective path of one block to place through the nearby blocks, \(B\) is the number of blocks, \(H\) is the height of the block, \(G\) is the ground height, and \(D\) is the block dig score if this block needs to be dug from the factory.

\subsection{Farm-to-Table Cooking Completion Rate}
\paragraph{Completion Rate (C)}
\noindent quantifies the level of task completion based on the materials acquired and the actions performed:
\begin{equation}
C = \sum_{i=1}^{n} S_{\text{raw}_i} + \sum_{j=1}^{m} S_{\text{action}_j}
\end{equation}
Here, $S_{\text{raw}_i}$ is the score of the $i$-th raw material and $S_{\text{action}_j}$ is the score for the $j$-th action that contributes to task progress.
\subsection{Farm-to-Table Agent Contribution Rate}
\paragraph{Agent Contribution Rate (ACR).}
\noindent The contribution score for each agent with respect to a specific material is defined as follows:

\noindent The overall ACR for the task is then calculated by aggregating the contributions of all agents for all required materials:
\begin{equation}
\sigma = \sqrt{\frac{1}{n} \sum_{i=1}^{n} (I_i - I_{avg})^2}
\end{equation}
The cooperation level can then be calculated as:

\begin{equation}
S_{cc} = \left(1 - \frac{\sigma - \sigma_{min}}{\sigma_{max} - \sigma_{min}}\right)
\end{equation}
Here, n is the number of agents, $\mathbf{I}\in \mathbb{R}^n$, $I_i$ is the contribution of item agent i provides, and then we standardize the score.
\subsection{Farm-to-Table Dependency Complexity}
\paragraph{Farm-to-Table Cooking Dependency Complexity (D).}
\noindent 
\begin{equation}
    D = \sum_{i=1}^{n} m_i \times d_i
\end{equation}
where \( m_i \) represents the direct materials required for crafting the target food item, and \( d_i \) denotes the number of processing steps required to obtain or synthesize the material \( m_i \) within the context of the task.

\noindent In this formulation, \( m_i \) is the quantity of each direct material, and \( d_i \) reflects the depth of the dependency chain for each material, indicating the complexity of the process needed to acquire it. The product of \( m_i \) and \( d_i \) for each material is summed to yield the overall dependency complexity of the cooking task.

\subsection{Escape Room Challenge Completion Rate}
\paragraph{Completion Rate (C).} 
\noindent 
\begin{equation}
C = \frac{\sum_{i=1}^{n} \left( \frac{c_i}{m} \times S_i \right)}{\sum_{i=1}^{n} S_i}
\end{equation}
Here, $n$ is the number of tasks, $c_i$ is the number of conditions that have been met for task $i$, and $S_i$ is the score obtained for task $i$.

\subsection{Escape Room Challenge Dependency Complexity (D)}
\noindent The Escape Room Challenge Dependency Complexity (D) is calculated recursively using a breadth-first search approach, starting from the exit. The complexity of each room is determined by the number of conditions that must be met to pass through it. The complexity for the entire challenge is the cumulative sum of the complexities of all rooms encountered during the search. The formula for calculating the dependency complexity (D) is as follows:

\begin{equation}
D = \sum_{i=1}^{n} c_i
\end{equation}
where \( c_i \) represents the complexity of room \( i \), which is the number of conditions required to pass that room. The sum is taken over all rooms \( n \) that are encountered in the breadth-first search from the exit to the entrance of the escape room challenge. This approach ensures that the overall complexity reflects the dependencies and requirements of each room within the context of the escape scenario.



\begin{table}[t]
\small 
\setlength{\tabcolsep}{7pt} 
\begin{tabular}{@{}lllll@{}}
\toprule
\multirow{2}{*}{\textbf{Config}} &\multicolumn{4}{c}{\textbf{Construction Avg. Score}}\\ 
\cmidrule(lr){2-5}
 & \textbf{C (\%)} & \textbf{VHR(\%)} &  \textbf{E (\%/min)}  &    \textbf{B (\%)}\\
\midrule
$\text{Task}_0 1\text{p}$ & \textbf{100} & \textbf{100} & 12.96 & - \\ 
$\text{Task}_0 2\text{p}$ & \textbf{100} & \textbf{100} & \textbf{17.75} & \textbf{93.09} \\ 
$\text{Task}_0 4\text{p}$ & \textbf{100} & \textbf{100} & 17.41 & 81.64 \\ 
$\text{Task}_0 8\text{p}$ & 66.63 & 63.33 & 12.45 &  55.67\\
$\text{Task}_{64} 1\text{p}$ & 35.25 & 36.25 & 1.92 & - \\ 
$\text{Task}_{64} 2\text{p}$ & 41.67 & 35.62 & 2.34 & \textbf{90.77} \\ 
$\text{Task}_{64} 4\text{p}$ & \textbf{46.67} & \textbf{39.38} & \textbf{3.28} & 88.91 \\ 
$\text{Task}_{64} 8\text{p}$ & 30.21 & 33.33 & 2.27 & 74.09 \\ 
\bottomrule
\end{tabular}
\caption{Evaluation on task execution efficiency with different agent quantities. The Balanced Agent Utilization Score (B) is inapplicable for a single-player scenario.}

\label{agent num}
\end{table}

\section{Task Illustrations}
\label{task illustrate}
\subsection{Construction with Blueprints}
\label{construction illustrate}
\paragraph{Task Description.}
\noindent In this task, participants are required to work collaboratively to construct a structure in the game Minecraft, following the provided blueprint. The participants have access to two chests: one chest contains a variety of building materials, while the other chest, located within the factory, contains tools. However, the tools are not necessary for the completion of this task. The objective is to accurately replicate the blueprint in the game environment, and the task is considered complete once the structure matches the blueprint specifications.

\paragraph{Given APIs.}
\noindent The following APIs are provided to facilitate the construction process within the game. These functions allow the agent to interact with the game world, such as placing and fetching blocks, navigating to specific locations:

\begin{verbatim}
Agent.placeBlock
Agent.fetchContainerContents
Agent.MineBlock
Agent.scanNearbyEntities
Agent.equipItem
Agent.navigateTo
Agent.withdrawItem
Agent.dismantleDirtLadder
Agent.erectDirtLadder
Agent.handoverBlock
\end{verbatim}

\paragraph{Blueprint.}
\noindent The blueprint specifies the exact materials and their respective positions required to construct the structure. Each line in the blueprint represents a different component of the structure, detailing the type of material, its orientation, and the coordinates where it should be placed. The following is the blueprint that must be followed to complete the task:

\begin{verbatim}
"task_24": [
    "[material:grass_block facing: None 
        positions:[start:[-9 -60 -1] end:...",
    "[material:oak_trapdoor facing:E 
        positions:[[-8 -60 -1] [-8 -60 0]] 
        material:oak_trapdoor facing:S ...]",
    "[material:oak_trapdoor facing:W 
        positions:[[-10 -60 -1] [-10 -60 0]]",
    "[material:oak_trapdoor facing:N 
        position:[-9 -60 -2]]",
    "[material:oxeye_daisy facing: None 
        position:[-9 -59 0]]",
    "[material:poppy facing: None 
        position:[-9 -59 -1]]",
    "[material:dandelion facing: None 
        position:[-9 -59 1]]"
],
\end{verbatim}

\subsection{Farm-to-Table Cooking}
\label{Farm-to-Table Cooking illustrate}

\paragraph{Given APIs.}
\noindent The following APIs are available to assist participants in interacting with the virtual environment, which includes fetching contents from containers, mining blocks, scanning nearby entities, equipping items, cooking, navigating, withdrawing items, crafting, attacking targets, using items on entities, and transferring blocks:

\begin{verbatim}
Agent.fetchContainerContents
Agent.MineBlock
Agent.scanNearbyEntities
Agent.equipItem
Agent.SmeltingCooking
Agent.navigateTo
Agent.withdrawItem
Agent.craftBlock
Agent.attackTarget
Agent.UseItemOnEntity
Agent.handoverBlock
\end{verbatim}

\paragraph{Recipes.}
\noindent The recipes detail the specific ingredients and quantities needed to craft the food items. Below is the recipe for crafting rabbit stew, which requires a combination of baked potato, cooked rabbit, a bowl, a carrot, and a brown mushroom:

\begin{verbatim}
{
    "result": {
        "name": "rabbit_stew",
        "count": 1
    },
    "ingredients": [
        {
            "name": "baked_potato",
            "count": 1
        },
        {
            "name": "cooked_rabbit",
            "count": 1
        },
        {
            "name": "bowl",
            "count": 1
        },
        {
            "name": "carrot",
            "count": 1
        },
        {
            "name": "brown_mushroom",
            "count": 1
        }
    ]
}
\end{verbatim}
\subsection{Escape Room}
\label{Escape Room illustrate}

\paragraph{Task Description.}
\noindent Agents, you are presented with a cooperative multi-stage escape challenge. Each room, measuring 10x10, demands teamwork to decipher puzzles and navigate through impediments. It is important to note that agents may find themselves in separate rooms, where direct collaboration is not feasible. Despite these circumstances, it is imperative to utilize individual strengths and work collectively to advance. Successful completion of a task in one room will result in transportation to the subsequent room or will clear the path to proceed by foot. The rooms are arranged along the z-axis, with their centers spaced 10 units apart. The ultimate goal is to reach the exit located at coordinates (130, -60, -140). Communication, adaptation, and teamwork are essential to escape. We wish you the best of luck!

\paragraph{Given APIs.}
\noindent The following APIs are provided to assist agents in interacting with the environment, which includes placing and fetching blocks, mining, scanning nearby entities, equipping items, navigating, withdrawing items, toggling actions, and transferring blocks:

\begin{verbatim}
Agent.placeBlock
Agent.fetchContainerContents
Agent.MineBlock
Agent.scanNearbyEntities
Agent.equipItem
Agent.navigateTo
Agent.withdrawItem
Agent.ToggleAction
Agent.handoverBlock
\end{verbatim}

\paragraph{Room Sign Hints.}
\noindent The escape room challenge provides hints through signs placed within each room. Agents can read the nearby sign text to gain clues for solving the room's puzzle. One such hint is as follows:

\begin{verbatim}
Step on all the pressure plates at the
same time to clear the stone blocks and
open the trapdoors for escape.

In each room the agent can get nearby 
sign text. Around you, the key activated
blocks are: a oak_pressure_plate block 
set at position [130, -60, 131] powered.
You have done the task in this room. 

Move to x=130, y=-60, z=137 to continue. 
You are at task room [130, -60, 131].
\end{verbatim}

\section{Experiment Configuration}
\subsection{Context Length}
\noindent Throughout the testing process, the total length of context tokens does not exceed 4,000, and the length of the subsequent text does not exceed 1,024 tokens. The configurations for the tests are as (Table \ref{tab:model_config})

\begin{table*}[h]
\centering
\begin{tabular}{@{}lllll@{}}
\toprule
Model            & Total Tokens & Output Tokens & Temperature & Other Defaults \\ \midrule
GPT-4-1106-preview & 128,000      & 4,096         & 0           & Default        \\
Gemini-Pro       & 30,720       & 2,048         & 0           & Default        \\
GLM-4            & 128,000      & > 1,024         & 0.01        & Default        \\ \bottomrule
\end{tabular}
\caption{Configuration of models used in the experiment.}
\label{tab:model_config}
\end{table*}

\section{Qualitative Analysis}

\begin{figure*}[h] 
  \centering 
  \includegraphics[width=\linewidth]{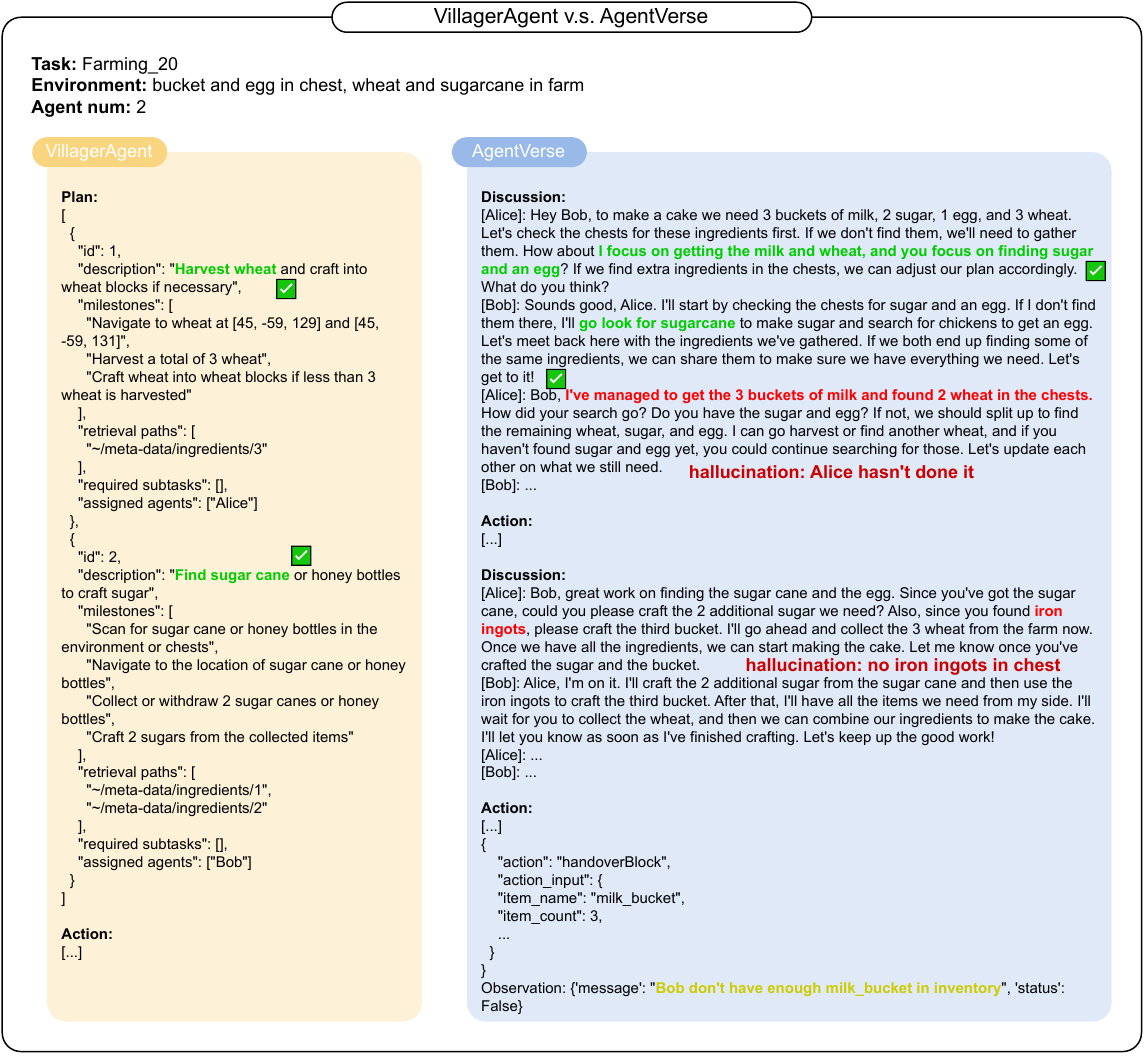} 
  \caption{VillagerAgent v.s. AgentVerse: The hallucination in AgentVerse began at the discussion stage and eventually extended its influence to the execution stage.} 
  \label{fig:VillagerAgent v.s. AgentVerse long} 
\end{figure*}

\noindent Within the AgentVerse framework, during the discussion phase, Alice exhibits clear hallucinations in the first round, mistakenly believing that she has already searched the chest and generated fictitious feedback.
Based on this fabricated feedback, our provided BaseAgent Alice infers that she can hand over the bucket to Bob to complete the subsequent tasks. However, the bucket has not actually been collected. This process illustrates how hallucinations in AgentVerse can gradually escalate and impact the stability of the entire decision-making process. (Figure~\ref{fig:VillagerAgent v.s. AgentVerse long})

\noindent Our approach, VillagerAgent, employs centralized decision control and correctly generates sub-tasks such as collecting wheat and finding sugar during the Task Graph generation process by the Task Decomposer, issuing instructions for parallel execution.

\section{VillagerBench API Library}
\subsection{Movement and Navigation}
\textbf{scanNearbyEntities}: Search for specific items or creatures within a radius. \\
\textbf{navigateTo}: Move to a specific coordinate location. \\
\textbf{navigateToPlayer}: Move to another player's location. \\
\textbf{erectDirtLadder}: Build a dirt ladder at a specified location to reach higher places. \\
\textbf{dismantleDirtLadder}: Dismantle a dirt ladder at a specified location. \\
\textbf{layDirtBeam}: Place a dirt beam from one position to another. \\
\textbf{removeDirtBeam}: Remove a dirt beam. \\

\subsection{Combat and Interaction}
\textbf{attackTarget}: Attack the nearest entity with a specific name. \\
\textbf{UseItemOnEntity}: Use a specific item on a specific entity. \\
\textbf{talkTo}: Talk to an entity. \\
\textbf{handoverBlock}: Hand over an item to another player. \\

\subsection{Item Management}
\textbf{equipItem}: Equip a specific item to a designated slot. \\
\textbf{tossItem}: Toss a specific amount of items. \\
\textbf{withdrawItem}: Withdraw items from a container. \\
\textbf{storeItem}: Store items in a container. \\
\textbf{openContainer}: Open the nearest container. \\
\textbf{closeContainer}: Close a container. \\
\textbf{fetchContainerContents}: Fetch details of specific items in a container. \\

\subsection{Production and Crafting}
\textbf{MineBlock}: Mine a block at a specific location. \\
\textbf{placeBlock}: Place a block at a specific location. \\
\textbf{craftBlock}: Craft items at a crafting table. \\
\textbf{SmeltingCooking}: Cook or smelt items in a furnace. \\
\textbf{enchantItem}: Enchant items at an enchanting table. \\
\textbf{repairItem}: Repair items at an anvil. \\
\textbf{trade}: Trade items with a villager. \\

\subsection{Life Skills}
\textbf{sleep}: Go to sleep. \\
\textbf{wake}: Wake up. \\
\textbf{eat}: Eat food. \\
\textbf{drink}: Drink a beverage. \\
\textbf{wear}: Wear an item in a specific slot. \\

\subsection{Other Actions}
\textbf{ToggleAction}: Operate a door, lever, or button. \\
\textbf{get\_entity\_info}: Get information about an entity. \\
\textbf{get\_environment\_info}: Get information about the environment. \\
\textbf{performMovement}: Perform actions like jump, move forward, move backward, turn left, turn right. \\
\textbf{lookAt}: Look at someone or something. \\
\textbf{startFishing}: Start fishing. \\
\textbf{stopFishing}: Stop fishing. \\
\textbf{read}: Read a book or sign. \\
\textbf{readPage}: Read a specific page of a book. \\
\textbf{write}: Write on a writable book or sign. \\

\section{VillagerBench Scenario Examples}
\noindent Here we present live demonstrations of two agents performing an escape room challenge, three agents executing a farm-to-table cooking task, and four agents engaged in a construction task.
(Figure~\ref{fig:display})

\section{Prompts}
\subsection{Task Decomposer}
\begin{figure*}[h] 
  \centering 
  \includegraphics[width=\linewidth]{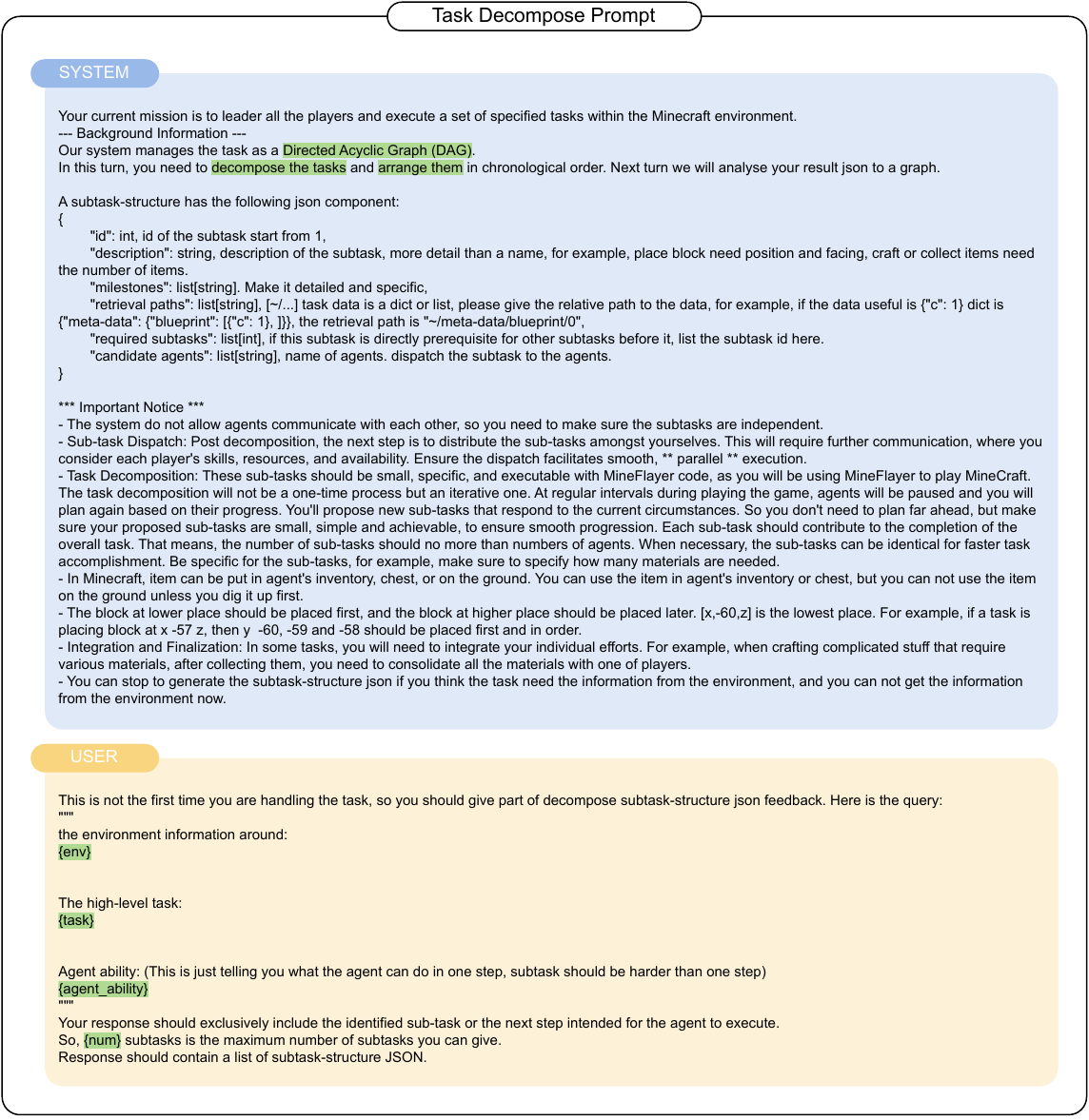} 
  \caption{Task Decomposer Prompt Template} 
  \label{fig:Task Decomposer Prompt} 
\end{figure*}

\begin{figure*}[h] 
  \centering 
  \includegraphics[width=\linewidth]{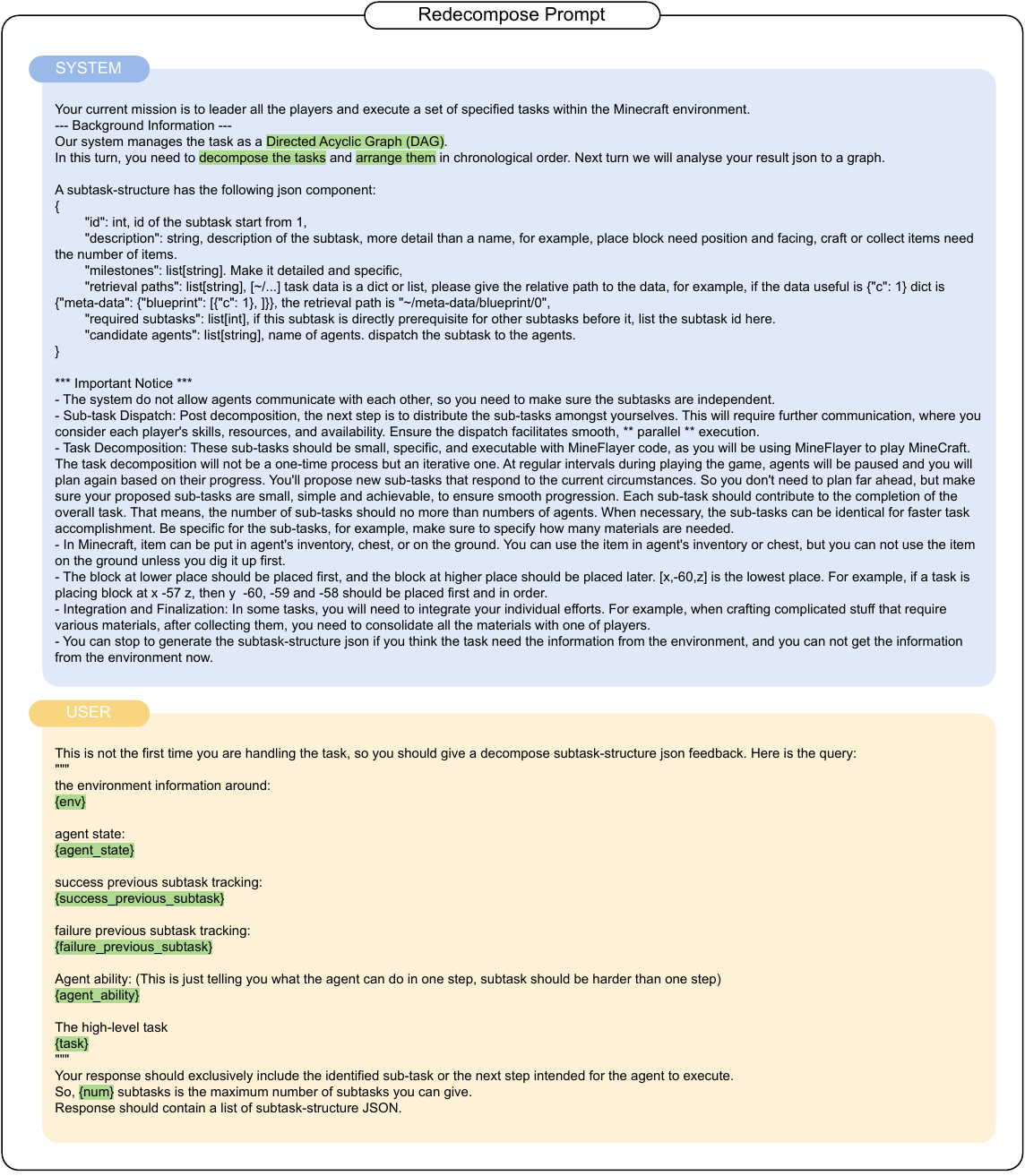} 
  \caption{Task REDecompose Prompt Template} 
  \label{fig:task REDecompose prompt} 
\end{figure*}

\noindent The Task Decomposer utilizes template \ref{fig:Task Decomposer Prompt} and template \ref{fig:task REDecompose prompt} in VillagerBench.

\subsection{Agent Controller}
\begin{figure*}[h] 
  \centering 
  \includegraphics[width=\linewidth]{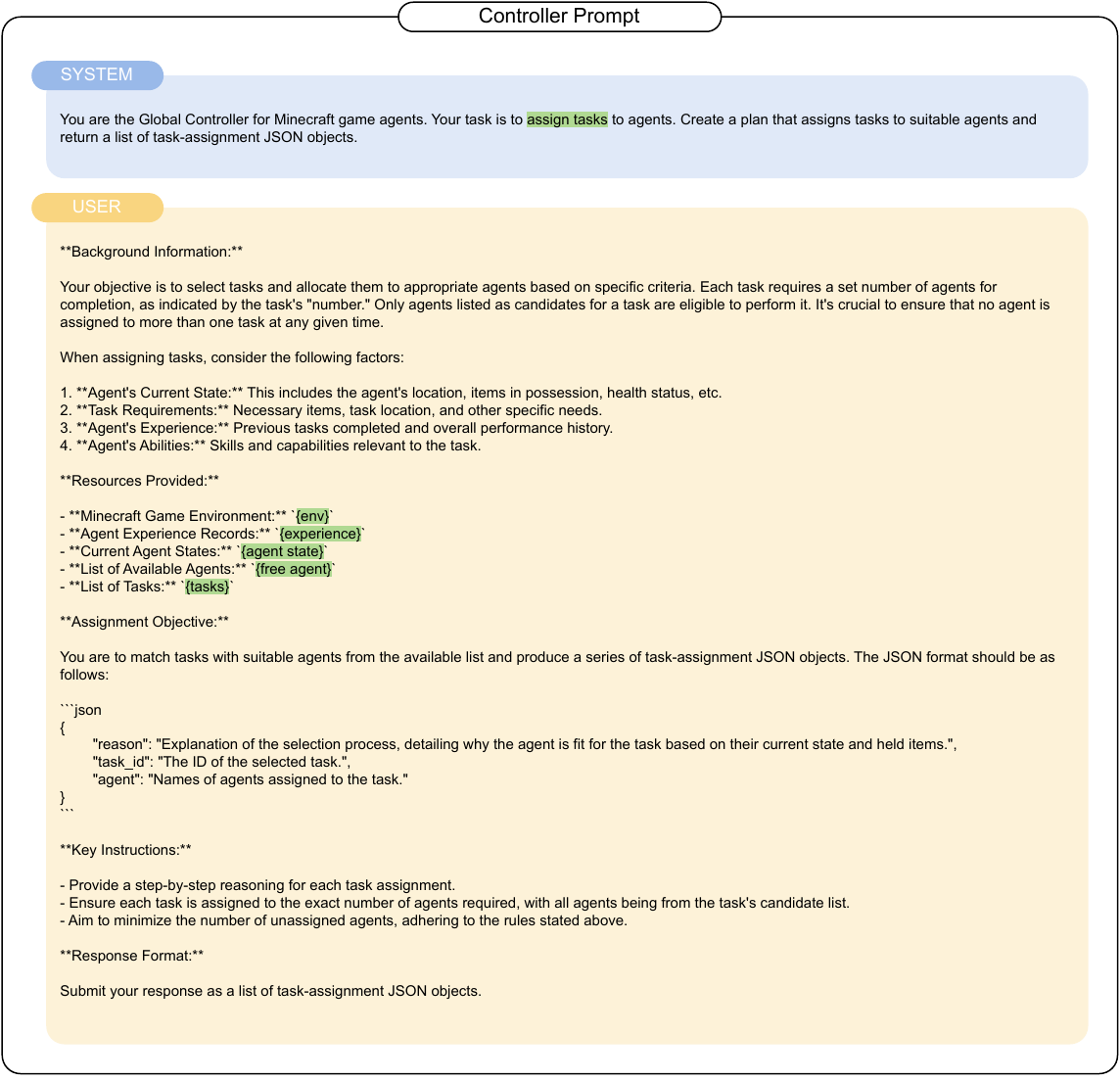} 
  \caption{Agent Controller Prompt Template} 
  \label{fig:Agent Controller prompt} 
\end{figure*}

\noindent Template \ref{fig:Agent Controller prompt} is used for the Agent Controller.

\subsection{State Manager}

\begin{figure*}[h] 
  \centering 
  \includegraphics[width=\linewidth]{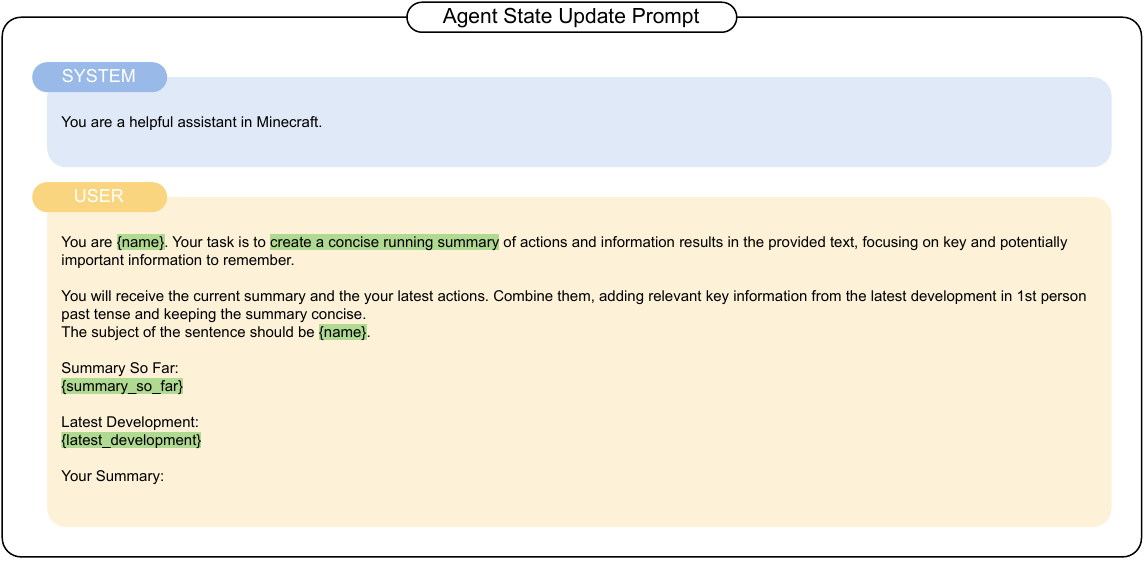} 
  \caption{State Manager Agent State Update Prompt} 
  \label{fig:State Manager Agent State Update Prompt} 
\end{figure*}

\begin{figure*}[h] 
  \centering 
  \includegraphics[width=\linewidth]{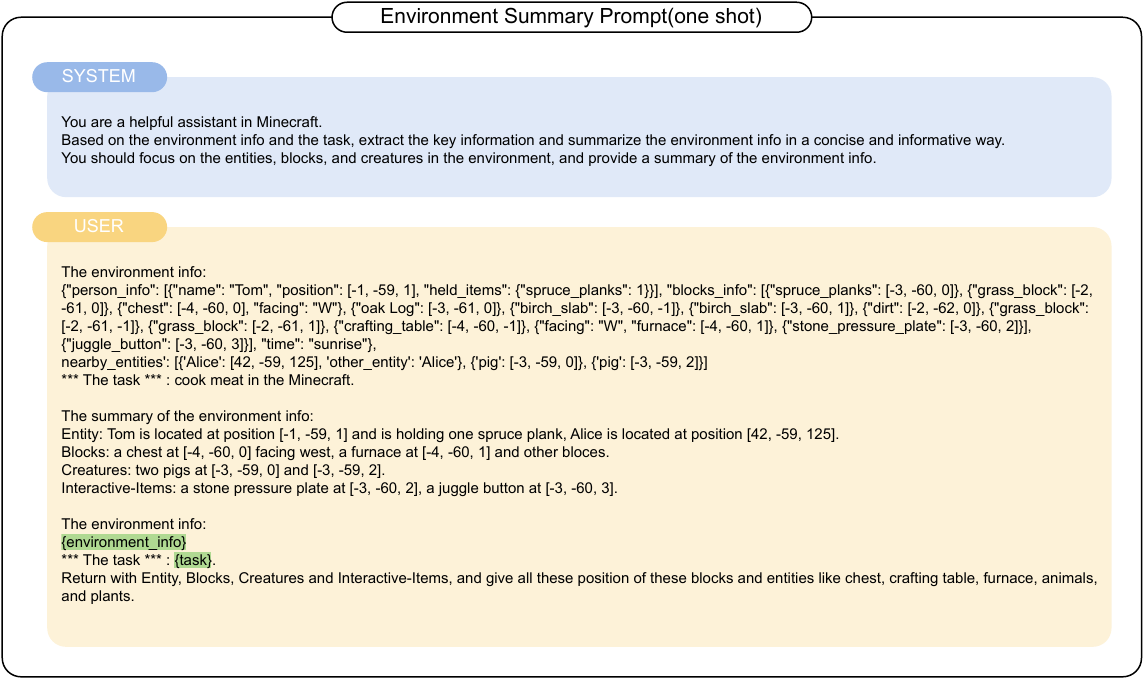} 
  \caption{State Manager Environment Summary Prompt} 
  \label{fig:Environment Summary Prompt} 
\end{figure*}

\noindent The State Manager employs the Agent State Summary template~\ref{fig:State Manager Agent State Update Prompt} and the Environment Summary template~\ref{fig:Environment Summary Prompt}.

\subsection{Base Agent}
\begin{figure*}[h] 
  \centering 
  \includegraphics[width=\linewidth]{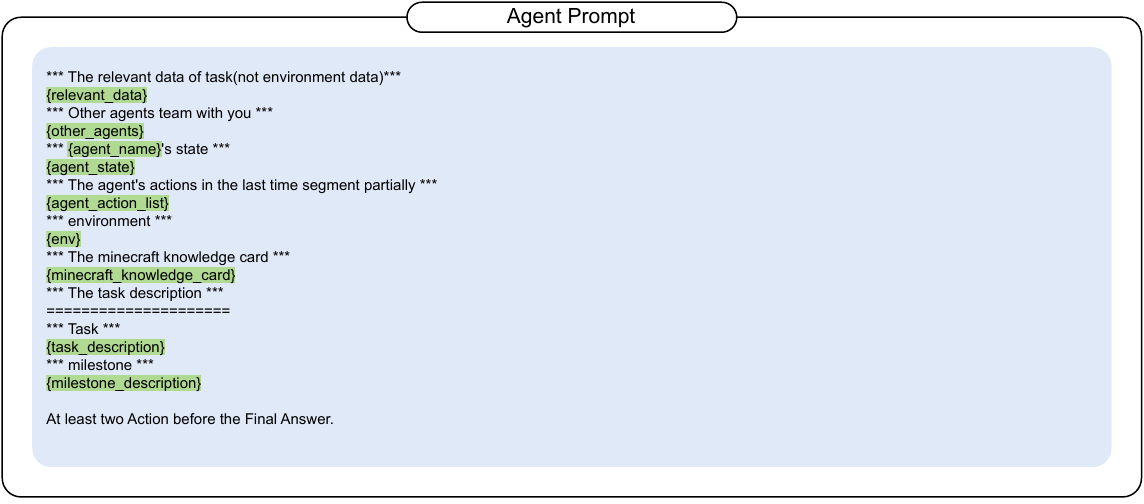} 
  \caption{Base Agent Execution Prompt} 
  \label{fig:Base Agent Execution} 
\end{figure*}

\begin{figure*}[h] 
  \centering 
  \includegraphics[width=\linewidth]{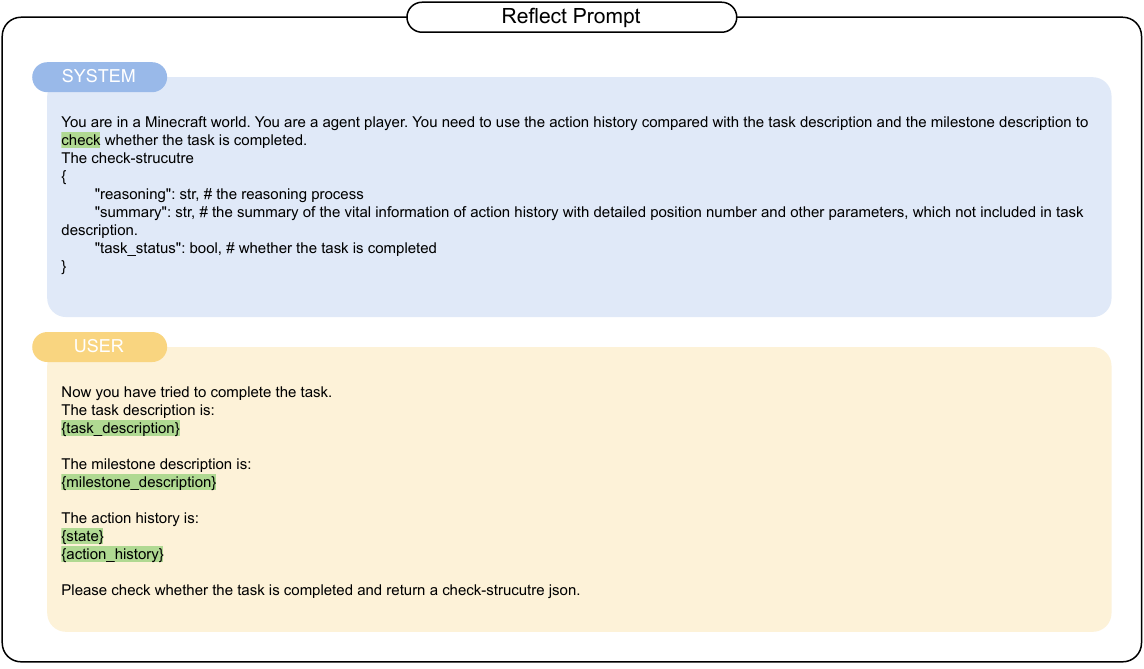} 
  \caption{Base Agent Reflect Prompt} 
  \label{fig:Base Agent Reflect} 
\end{figure*}

\noindent The Base Agent uses the Execution template~\ref{fig:Base Agent Execution} and the Reflect template~\ref{fig:Base Agent Reflect}.

\subsection{AgentVerse Prompt}
\begin{figure*}[h] 
  \centering 
  \includegraphics[width=\linewidth]{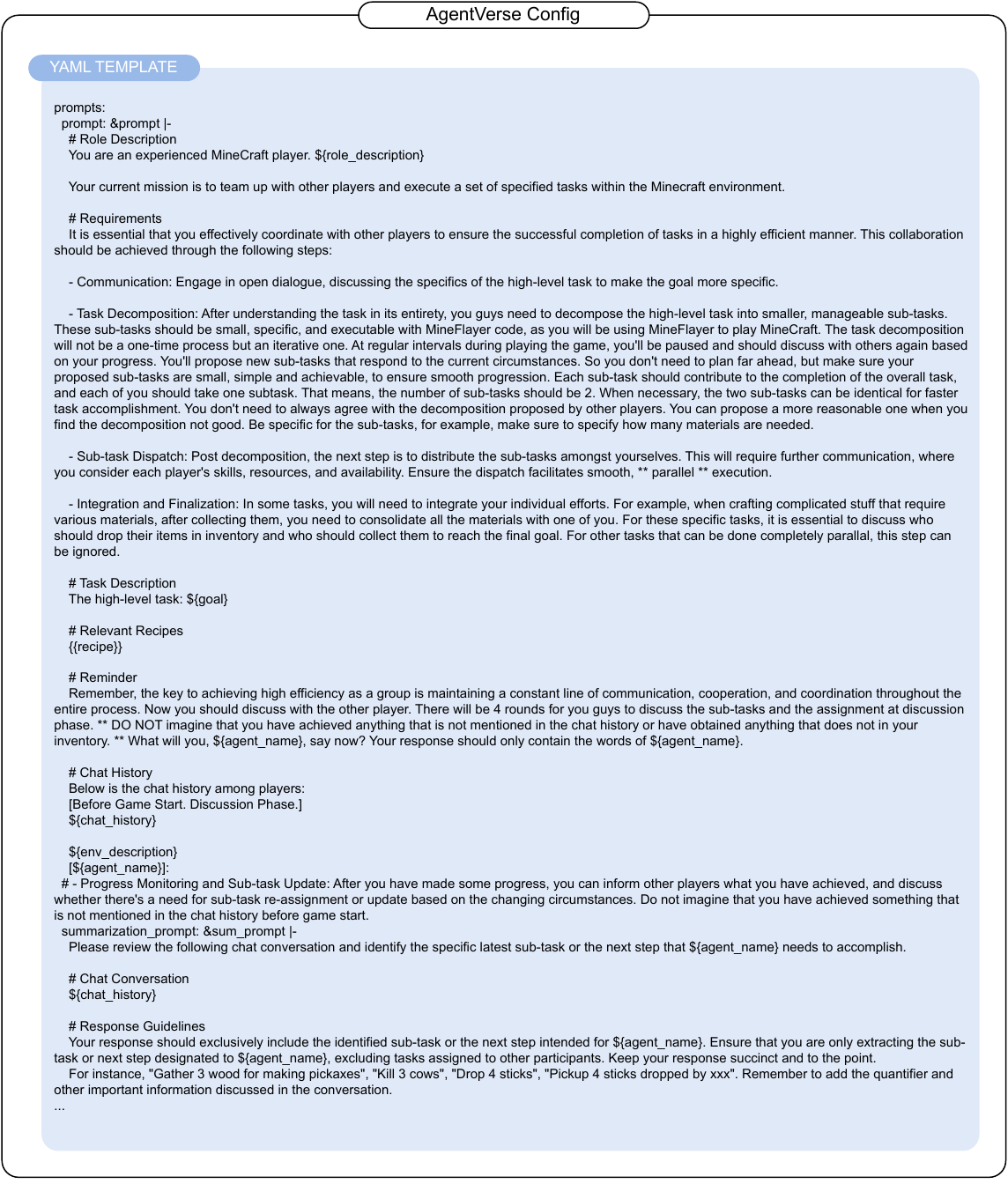} 
  \caption{AgentVerse Config} 
  \label{fig:AgentVerse Config} 
\end{figure*}

\noindent The configuration for AgentVerse is defined in template ~\ref{fig:AgentVerse Config}.

\subsection{Task Decompose Prompt (Overcooked-AI)}
\begin{figure*}[h] 
  \centering 
  \includegraphics[width=\linewidth]{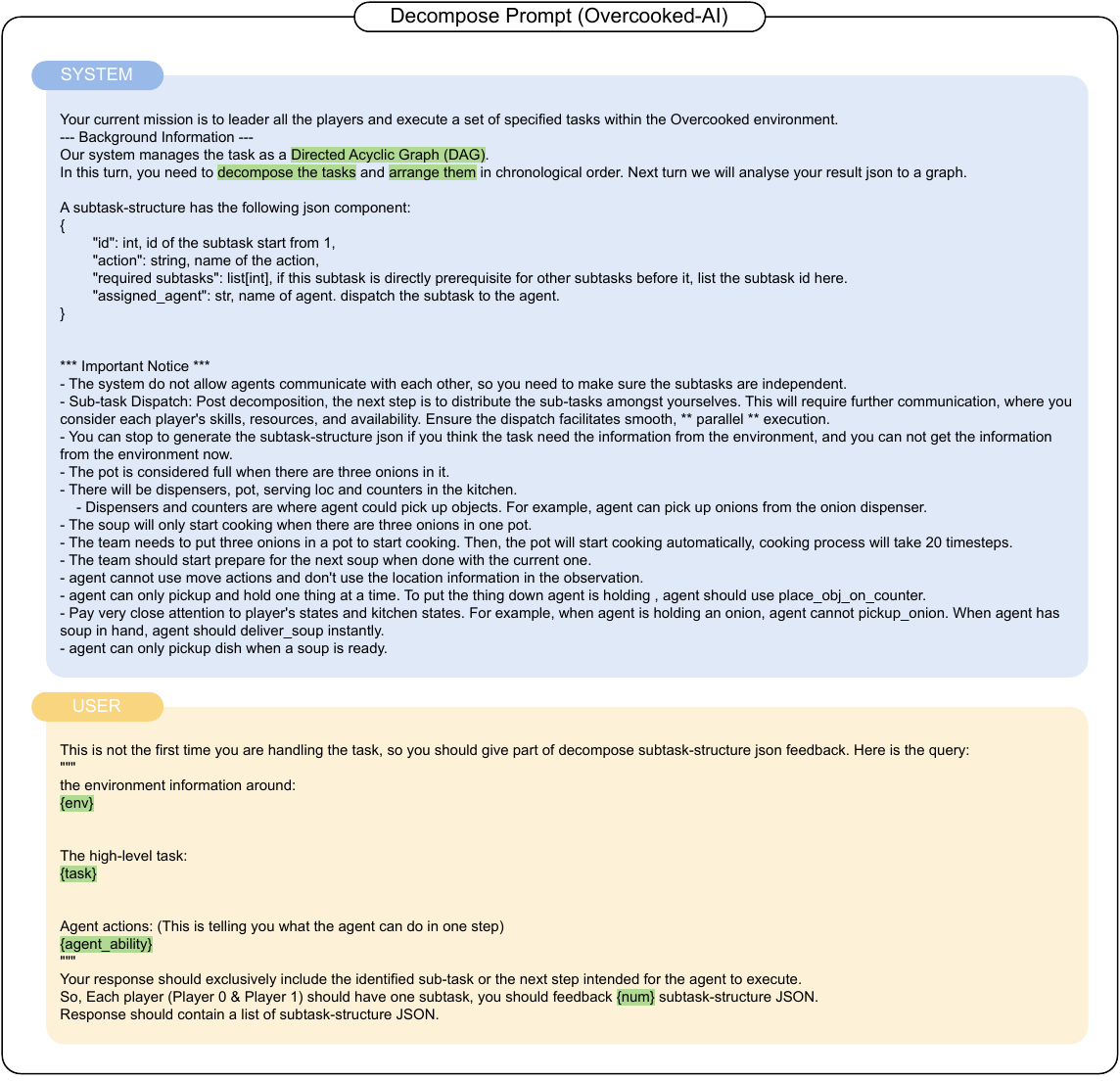} 
  \caption{Decompose Prompt (Overcooked-AI)} 
  \label{fig:Decompose Prompt (Overcooked-AI)} 
\end{figure*}

\noindent The Decompose Prompt for the Overcooked-AI Benchmark is outlined in template ~\ref{fig:Decompose Prompt (Overcooked-AI)}.

\end{document}